\RequirePackage[figuresleft]{rotating}
\PassOptionsToPackage{table}{xcolor}
\documentclass[acmsmall,nonacm]{acmart}

\usepackage{booktabs}
\usepackage{array}
\usepackage{tabularx}
\usepackage{colortbl}
\usepackage{adjustbox}
\usepackage{tikz}
\usepackage{algorithm}
\usepackage{algorithmic}
\usepackage{listings}
\usetikzlibrary{positioning,calc}
\definecolor{cTag}{RGB}{150,20,110}
\definecolor{cEnv}{RGB}{176,96,16}
\definecolor{cKw}{RGB}{20,80,180}
\definecolor{cStr}{RGB}{20,130,70}
\definecolor{cCom}{RGB}{125,125,135}
\definecolor{cBg}{RGB}{248,248,250}
\lstdefinestyle{prompt}{%
  alsoletter={_},%
  literate=%
    {[context]}{{\color{cTag}\bfseries[context]}}{9}%
    {[conditioning]}{{\color{cTag}\bfseries[conditioning]}}{14}%
    {[instruction]}{{\color{cTag}\bfseries[instruction]}}{13}%
    {[format]}{{\color{cTag}\bfseries[format]}}{8},%
  emph={CANDIDATE,END_CANDIDATE,PROMPT,SENT,SAMPLED,OUTPUT,VALIDATE,REPAIR,REPAIRED,EVALUATE},%
  emphstyle={\color{cEnv}\bfseries}}

\begin{document}

\title{From Hand-Crafted to LLM-Based Variation Operators in Metaheuristics: A Tutorial}
\author{Camilo Chac\'on Sartori}
\affiliation{\institution{Apeiron Intelligence}\city{Barcelona}\country{Spain}}
\affiliation{\institution{Artificial Intelligence Research Institute (IIIA-CSIC)}\city{Bellaterra}\country{Spain}}
\email{camilo.chacon@apeironagents.tech}

\author{Guillem Rodr\'iguez-Corominas}
\affiliation{\institution{Artificial Intelligence Research Institute (IIIA-CSIC)}\city{Bellaterra}\country{Spain}}
\affiliation{\institution{Universitat Polit\`ecnica de Catalunya (UPC)}\city{Barcelona}\country{Spain}}
\email{guillem.rodriguez.corominas@upc.edu}

\author{Christian Blum}
\affiliation{\institution{Artificial Intelligence Research Institute (IIIA-CSIC)}\city{Bellaterra}\country{Spain}}
\email{christian.blum@iiia.csic.es}

\renewcommand{\shortauthors}{Chac\'on Sartori, Rodr\'iguez Corominas \& Blum}

\begin{abstract}
Large language models (LLMs) are increasingly being employed as variation operators in metaheuristics, generating or modifying candidate solutions, heuristics, or programs inside iterative search loops. This shift reframes variation as a model call conditioned on different types of information. We introduce an operator-level framework with two descriptors: (1) the type of prompt-conditioning information at variation time (\texttt{Numeric}, \texttt{Symbolic}, \texttt{Linguistic}), and (2) artifact persistence, identifying what survives the model call (\texttt{Transient}, \texttt{Amortized}, \texttt{Transfer}). The tutorial shows how to classify, build, and select these operators through a worked build template, a method survey, an evidence table, and a cost-aware decision guide.
\end{abstract}


    \maketitle
\raggedbottom
\hfuzz=2pt
\hbadness=1200
\setlength{\emergencystretch}{1.2em}

\begin{center}
\small
\textbf{Paper website:} \url{https://camilochs.github.io/semantic-turn-metaheuristics}
\end{center}

\section{Introduction}
\label{sec:intro}

Metaheuristic search is shaped by variation operators such as local search moves, mutation, recombination, or edit rules that generate new candidate solutions. In classical metaheuristics, such operators are usually hand-designed. Machine-learning integrations from past years often assisted the search process of metaheuristics by predicting objective function values (solution scores), selecting operators, or warm-starting solvers, while variation operators themselves remained fixed by the designer. A growing class of LLM-based methods changes this paradigm~\cite{romeraparedes2024funsearch,liu2024eoh,ye2024reevo,vanstein2025llamea}: an LLM generates or edits candidate solutions, heuristics, or programs directly at the variation step, inside an iterative search loop. We call this shift a \emph{semantic turn}: variation becomes a model call conditioned on specific types of information at \emph{variation time}, i.e., the time at which the variation operator is executed.

This semantic shift raises practical questions that existing taxonomies of variation operators do not address. Consider a practitioner developing an LLM-assisted Traveling Salesman Problem (TSP) metaheuristic search procedure: should the prompt include (a)~only performance scores of candidate heuristics, (b)~the current heuristic as executable code, or (c)~a natural-language reflection on why the heuristic stalls? These choices can shift costs substantially across validation, evaluation, and LLM inference, so the right choice is domain- and budget-dependent. The literature contains many such methods, but lacks a shared operational framework for navigating among them.

This tutorial is written for researchers and practitioners in optimization---especially for those developing iterative, stochastic search processes such as metaheuristics---who want to beneficially incorporate LLMs into their methods. We assume familiarity with basic metaheuristic loops, but not with transformer architectures or LLM internals; Section~\ref{sec:background} provides the LLM background needed here. We focus on variation operators conditioned through text prompts. Fine-tuned or adapter-based models remain in scope when the variation step is still prompt-conditioned. In those cases, we classify the prompt channel in the same way. By contrast, operators whose conditioning is encoded only in the model parameters or supplied through non-text inputs such as images fall outside our scope and are discussed as open directions in Section~\ref{sec:open}.

\paragraph{The organizing lens.}
We view existing applications through one structural question: \emph{on what type of problem structure is the operator conditioned?} We distinguish three conditioning channels---\texttt{Numeric}, \texttt{Symbolic}, and \texttt{Linguistic}---and pair them with a second descriptor, \emph{artifact persistence}, which describes how the emitted artifact is used within the search: consumed inside the current search loop, reused after offline design, or re-bound to a new instance family or domain. Together, the two descriptors form an operator-level taxonomy for practice: they allow us to classify methods, guide the construction of validated prompt-conditioned variation loops, and support family-level choices under evaluator, budget, and deployment constraints.\footnote{Throughout, the lens is descriptive rather than predictive: it organizes the design space, but does not claim a universal law linking more explicit problem representations in the prompt to better optimization performance. Whether such structure helps remains an empirical, task-dependent question, which we return to in Section~\ref{sec:open}.}

The lens is a complementary teaching spine: Table~\ref{tab:positioning} contrasts it with surveys organized by model role, the direction of interaction between optimization algorithms and LLMs, integration stage, or systematic-review metadata such as task, architecture, dataset, and application. Readers familiar with those surveys should find here a practical route through the operator-design problem: identify the load-bearing prompt content, implement a validated variation loop, and decide when the LLM belongs in the search pipeline.

\begin{table}[!t]
\rowcolors{2}{white}{gray!15}
\caption{Complementary positioning of this tutorial relative to recent surveys.}
\label{tab:positioning}
\footnotesize
\setlength{\tabcolsep}{3pt}
\renewcommand{\arraystretch}{1.15}
\begin{tabularx}{\linewidth}{@{}>{\raggedright\arraybackslash}p{2.5cm}
                  >{\raggedright\arraybackslash}p{1.9cm}
                  >{\raggedright\arraybackslash}X
                  >{\raggedright\arraybackslash}X@{}}
\toprule
\textbf{Work} & \textbf{Type of work} & \textbf{Organizing axis} & \textbf{After reading, you can}\dots \\
\midrule
LLM4AD~\cite{liu2025llm4ad} (CSUR) & survey & the LLM's \emph{role}: optimizer / predictor / extractor / designer & locate a method by the role the LLM plays \\
Da Ros et al.~\cite{daros2026llm4co} (CSUR) & systematic review (PRISMA) & task, architecture, dataset, application & find the census of LLM-for-CO (combinatorial optimization) studies \\
Wu et al.~\cite{wu2024ecera} & survey + roadmap & direction of enhancement: LLM-enhanced EA vs.\ EA-enhanced LLM & see how evolutionary computation (EC) and LLMs enhance each other \\
Zhang et al.~\cite{zhang2025llmevopt} & survey & integration stage: modeling $\to$ solving & trace the modeling-to-solving pipeline \\
\textbf{This tutorial} & \textbf{tutorial} & what the operator is \emph{conditioned on}: \texttt{Numeric} / \texttt{Symbolic} / \texttt{Linguistic} & \textbf{build, classify,} and \textbf{choose} an operator \\
\bottomrule
\end{tabularx}
\end{table}

Our contribution is threefold. First, we introduce a two-descriptor framework for LLM variation operators: dominant conditioning channel and artifact persistence. Second, we make the framework operational through a reusable build template with parsing, validation, and bounded repair. Third, we synthesize representative methods through that framework and give a cost-aware decision guide for choosing among them.

The paper is organized to mirror this workflow. Section~\ref{sec:background} shows how an LLM call can be read as a stochastic variation operator. Section~\ref{sec:lens} defines the conditioning and persistence descriptors, grounds the term ``semantic'' in the classical syntax/semantics distinction, and gives the placement rule. Section~\ref{sec:build} teaches the mechanics of building an operator, from the prompt template to the validator and the repair loop. Section~\ref{sec:tour} places representative methods on the lens, and Section~\ref{sec:choose} turns that map into a decision guide. Sections~\ref{sec:crosscut} and~\ref{sec:open} collect the engineering risks and open research directions, including the reproducibility and reporting discipline needed for credible claims. The appendices provide a copyable prompt template, a Python reference loop, a channel-ablation audit, and an extended coverage map. To support practical adoption and reproducibility, we provide a repository\footnote{\href{https://camilochs.github.io/semantic-turn-metaheuristics}{https://camilochs.github.io/semantic-turn-metaheuristics}} with the materials developed for this tutorial, including prompt templates, reference code, and other supplementary resources.

\section{Background and preliminaries}
\label{sec:background}
This section introduces the background needed to view prompt-conditioned LLM operators as variation mechanisms: the structure of metaheuristic search loops, the mechanics of LLMs as conditional samplers, the proposal-distribution framing that connects the two, and related research fields that motivate the boundary of the classification framework that we introduce (later called \emph{the lens}). In this context, note that---in accordance with LLM notations and expression---we often call the output of an LLM-based variation operator a proposal.

\paragraph{The metaheuristic search loop.}
A metaheuristic maintains a search state---an incumbent solution, a population of solutions, an archive, or a set of candidate artifacts---and repeatedly proposes, evaluates, and selects among alternatives~\cite{talbi2009metaheuristics,gendreau2019handbook}. The artifact being varied may be a solution, a move rule, a heuristic, or a program, depending on the level at which search is being performed. We use a single-artifact schematic only to mark the proposal step that an LLM may replace:
{\setlength{\topsep}{9pt}%
\begin{center}
{\setlength{\fboxsep}{3pt}%
\fbox{\begin{minipage}{0.88\linewidth}\ttfamily\footnotesize
x $\leftarrow$ initial\_solution()\\
best\_seen $\leftarrow$ x\\
while not stop\_condition:\\
\hspace*{2em}x' $\leftarrow$ {\normalfont\textsc{Vary}}(x)\hfill\# the variation operator\\
\hspace*{2em}if f(x') better than f(best\_seen): best\_seen $\leftarrow$ x'\\
\hspace*{2em}if {\normalfont\textsc{Accept}}(x', x): x $\leftarrow$ x'\\
return best\_seen
\end{minipage}}}
\end{center}}
Here \textsc{Accept} denotes the selection rule, which may be greedy or probabilistic; \texttt{best\_seen} is an elitist record tracked separately because the current state can temporarily be worse than the best candidate found. The shift studied here replaces a hand-coded \textsc{Vary} --- for example, bit-flip and 2-opt neighborhoods in stochastic local search~\cite{hoos2004sls}, tabu-list restrictions~\cite{glover1986tabu}, or Ant Colony Optimization construction rules~\cite{dorigo2004aco} --- with a prompt-conditioned LLM call. In that case, parsing and bounded repair occur logically between \textsc{Vary} and \textsc{Accept}: a candidate that cannot be repaired is rejected before the acceptance rule evaluates it. Because such calls are expensive in latency, token use, and API or deployment cost relative to classical moves, the call budget shapes when and how often the model should be invoked (Sections~\ref{sec:build} and~\ref{sec:choose}). Algorithm~\ref{alg:llmop} later formalizes this schematic with the notation used in the build section.

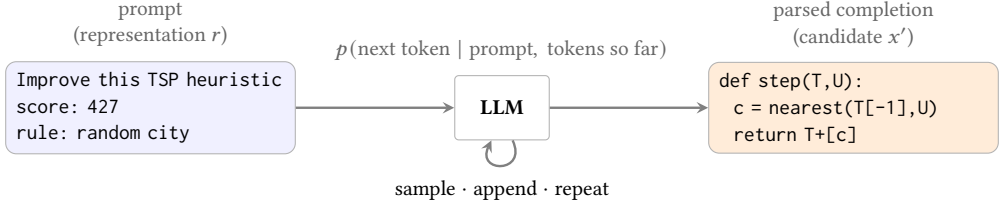
\begin{figure}[!t]
\centering
\begin{tikzpicture}[font=\footnotesize,>=stealth,
  box/.style={draw=black!30,rounded corners,align=left,inner sep=4pt,font=\footnotesize},
  tinylabel/.style={font=\footnotesize,text=black!60}]
 \node[box,fill=blue!6,text width=3.55cm] (p) at (0,0)
   {\texttt{Improve this TSP heuristic}\\
    \texttt{score: 427}\\
    \texttt{rule: random city}};
 \node[tinylabel,above=2pt of p,align=center] {prompt\\(representation $r$)};
 \node[draw=black!30,rounded corners=1.5pt,fill=white,minimum width=1.25cm,
       minimum height=0.85cm,align=center,font=\footnotesize\bfseries] (m) at (4.65,0) {LLM};
 \node[tinylabel,above=2pt of m,align=center]
   {$p(\text{next token}\mid\text{prompt},\ \text{tokens so far})$};
 \draw[->,black!50,line width=0.9pt] (p) -- (m);
 \node[box,fill=orange!15,text width=3.55cm] (c) at (9.3,0)
   {\texttt{def step(T,U):}\\
    \texttt{\ \ c = nearest(T[-1],U)}\\
    \texttt{\ \ return T+[c]}};
 \node[tinylabel,above=2pt of c,align=center] {parsed completion\\(candidate $x'$)};
 \draw[->,black!50,line width=0.9pt] (m) -- (c);
 \draw[->,black!50,line width=0.9pt] ([xshift=4pt]m.south) to[out=-65,in=-115,looseness=4.4]
   node[midway,below=2pt,font=\footnotesize,text=black]{sample $\cdot$ append $\cdot$ repeat}
   ([xshift=-4pt]m.south);
\end{tikzpicture}
\caption{Token-level view of an LLM call. A prompt containing the representation $r$ is converted into a completion by repeatedly sampling a next token conditioned on the prompt and the tokens already generated. The completed text is then parsed as a candidate artifact $x'$.}
\Description{A prompt representation is tokenized, passed through an LLM, decoded into a completion, and parsed into a candidate artifact.}
\label{fig:sampler}
\end{figure}

\paragraph{An LLM is a conditional sampler.}
Several facts about LLMs are sufficient for this tutorial. A \emph{prompt} is the text fed to the model. Text is represented as \emph{tokens}, discrete units from the model's fixed vocabulary; at each generation step the model samples one token from a probability distribution over that vocabulary, extending the sequence until a stopping criterion is met (Fig.~\ref{fig:sampler}). A \emph{completion} is the decoded sequence returned by the model.\footnote{Many APIs also expose structured-output schemas, tool/function calling, or grammar-constrained decoding, which constrain the form of a completion before parsing; in this tutorial they are validation aids and generation-interface choices, not new conditioning channels~\cite{geng2023grammar,willard2023efficient}.} \emph{Temperature} controls randomness by flattening or sharpening this next-token distribution (higher is more diverse, lower more stable). Top-$k$ and top-$p$ are common decoding filters over the next-token choices: top-$k$ restricts sampling to the $k$ most likely next tokens, while top-$p$ restricts it to the smallest set of likely tokens whose cumulative probability mass reaches $p$. These decoding controls reshape the induced proposal distribution, analogous to how mutation rates or step sizes tune a classical stochastic operator.

\emph{In-context} conditioning embeds task context, examples, or both in the prompt to induce task-specific behaviour without changing the model's weights\footnote{Those conditional probabilities come from pretraining: the model's weights encode regularities learned from large corpora of text and code, which shape the next-token distribution at inference time.}~\cite{brown2020gpt3}. \emph{Few-shot} conditioning is the special case where example input--output pairs are included; \emph{zero-shot} conditioning provides only a task description or search-state representation, with no exemplar completions. One useful view of in-context learning is that the prompt acts as evidence about the task the model is being asked to perform, so examples and surrounding context can shift which completions are likely under the fixed weights~\cite{xie2022iclbayes}.\footnote{Prompts can also ask the model to expose intermediate reasoning before producing the final artifact; the artifact tokens are then sampled conditional on that stated rationale, echoing chain-of-thought (\emph{CoT}) prompting~\cite{wei2022cot}.}

Prompt phrasing, ordering, and delimiters create format sensitivity: because the model predicts each next token conditioned on the exact preceding text, even substantively equivalent rephrasings can change which continuations are likely. Long prompts also add a placement issue: models often use information at the beginning or end of context more reliably than information buried in the middle~\cite{liu2023lostmiddle}. For our purposes, the prompt is the designer's main interface to the sampler (Section~\ref{sec:lens}): changing what it contains reshapes the distribution of all subsequent token choices.

The previous points describe a single completion; across a search run, feedback must be carried from one call to the next. At the API level, each LLM call is stateless: the model's weights do not change between calls.\footnote{Session-based APIs may accumulate conversational context server-side, but the reliable design assumption is that adaptation across search steps must be written into successive prompts.} Each call also has a finite \emph{context window}, the maximum combined length of the prompt and completion that the model can process. Accumulated feedback must therefore be summarized or pruned when it exceeds that budget, introducing compression losses.

\begin{figure}[!t]
\centering
\begin{tikzpicture}[font=\footnotesize,
  distlabel/.style={font=\footnotesize,text=black},
  sublabel/.style={font=\footnotesize,text=black!60}]
 \begin{scope}
   \draw[draw=blue!55!black,line width=0.9pt,fill=blue!12]
     plot[smooth,domain=-1.85:1.85,samples=60] (\x,{1.25*exp(-2.2*\x*\x)}) -- cycle;
   \draw[->,black!50,line width=0.9pt] (-2.0,0) -- (2.2,0) node[right,black]{$x'$};
   \fill[black] (0,0) circle (1.3pt) node[below=1pt,black]{$x$};
	   \node[distlabel] at (0,1.7) {$q(x'\mid \text{state})$};
	   \node[sublabel] at (0,-0.72) {\itshape classical (fixed kernel)};
 \end{scope}
 \begin{scope}[xshift=5.8cm]
   \draw[draw=orange!75!black,line width=0.9pt,fill=orange!14]
     plot[smooth,domain=-1.95:1.95,samples=80]
       (\x,{0.5*exp(-3.5*(\x+0.75)*(\x+0.75))+1.2*exp(-2.8*(\x-0.7)*(\x-0.7))}) -- cycle;
   \draw[->,black!50,line width=0.9pt] (-2.0,0) -- (2.2,0) node[right,black]{$x'$};
   \fill[black] (0,0) circle (1.3pt) node[below=1pt,black]{$x$};
	   \node[distlabel] at (0,1.7) {$q(x'\mid \text{prompt } r)$};
	   \node[sublabel] at (0,-0.72) {\itshape LLM (explicit prompt $r$)};
 \end{scope}
\end{tikzpicture}
\caption{The operator as a proposal distribution $q(x'\mid r)$. Left: a classical fixed kernel produces a narrow proposal centred on the current state $x$. Right: an LLM prompt can induce a different, sometimes broader distribution over candidate artifacts.}
\Description{Two schematic proposal distributions compare a classical neighborhood around a current state with an LLM-induced distribution shaped by a prompt representation.}
\label{fig:proposal}
\end{figure}
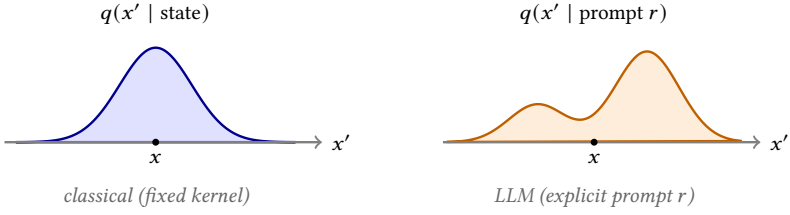

For search, the model's stochasticity --- diversity across completions on repeated calls --- is an asset: it drives exploration in the role of classical perturbation. This stochastic decoder becomes useful only after it is embedded in an operator. An LLM variation operator can be viewed as a \textbf{proposal distribution} $q(x'\mid r)$ that maps a representation $r$ of the problem and current search state to a distribution over candidate artifacts $x'$ (Fig.~\ref{fig:proposal}) --- a notion already familiar from stochastic local search. In the schematic baseline, a classical operator applies a fixed or hand-designed move, neighborhood, mutation, or construction rule. In an LLM operator, the central design object is instead $r$: the explicit representation placed in the prompt at generation time. Changing $r$ changes the induced proposal distribution and, therefore, the properties of the resulting artifacts ---including whether they can be parsed, validated, executed externally, or compared.

Strictly, the model samples token sequences, not solutions or programs. Because next-token likelihood is not the same as domain feasibility or executable correctness, outputs can be malformed, infeasible, or non-executable. Parsing, validation, and \emph{bounded repair} (a limited number of repair attempts after invalid output) map those sequences to candidate artifacts, rejecting unrepaired outputs before selection. Thus $q(x'\mid r)$ denotes the artifact-level proposal distribution induced by the whole variation module, not just the raw next-token distribution of the LLM. Unless stated otherwise, the notation suppresses fixed choices such as model version, decoding settings, parser, validator, and repair policy. Both classical and LLM operators are treated here as stochastic proposal mechanisms; nothing in this framing requires the model to ``understand'' anything.

The notation is useful because it names measurable design effects. Changing $r$ can alter the probability that raw completions parse and validate, the number of repair attempts per accepted artifact, the locality or diversity of proposals, and the probability mass near feasible high-quality artifacts. High invalid-output or repair rates are therefore diagnostic: the model's raw completions have poor overlap with the feasible artifact space, so $r$ is not steering the variation module toward valid candidates.

From an optimization perspective, this recalls Estimation of Distribution Algorithms (EDAs)~\cite{larranaga2002eda}, which explicitly learn and sample candidate distributions; here the artifact-level distribution is induced implicitly by prompt content, decoding, parsing, validation, and repair. This analogy is only a bridge for optimization readers: LLM operators do not inherit EDA modeling assumptions or formal properties.

\paragraph{Boundaries with related literature.}
Prompt-conditioned variation is a continuation of a long-running integration of machine learning into metaheuristics, with a deep-learning wave accelerating over the past decade, including \emph{ML into metaheuristics}~\cite{talbi2021ml,karimimamaghan2022ml,dacostaoliveira2023ml}, \emph{learnheuristics}~\cite{calvet2017learnheuristics}, \emph{ML for combinatorial optimization}~\cite{bengio2021ml}, and hybrid metaheuristics~\cite{blum2011hybrid}. Some LLM-hybrid work remains assistive: it extracts instance patterns that inform a conventional metaheuristic rather than generating the variation itself~\cite{chaconsartori2025joinforces}. Here, we focus on the point where the prompt-conditioned model call becomes part of the variation operator itself. The boundaries below clarify how that prompt interface differs from concepts in related fields.

Automated algorithm design is the closest relative to LLM-based automatic heuristic design. In this part of the literature, the model proposes or revises the heuristic, operator, or program that will later be run by the solver. Two boundary cases delimit the space. Algorithm configuration~\cite{lopezibanez2016irace,hutter2011smac} searches over parameters of a fixed algorithm structure; an LLM operator reduces to configuration when the prompt carries parameter proposals but does not change the operator family (the underlying move class, heuristic template, or solver component being tuned). Algorithm selection~\cite{rice1976algselection,kotthoff2014algsel} chooses among a predefined portfolio using problem or runtime features; an LLM-based routing method is a prompt-conditioned instantiation of selection when it selects among fixed operators. Code-emitting operators that synthesize new heuristics or programs sit at the design end of the spectrum.

Hyper-heuristics~\cite{burke2013hyper,burke2019hyperrevisited} also operate on a higher level, selecting or generating heuristics and, in cross-domain forms, transferring control strategies across problem families. LLM-based variants inherit that ambition, but add prompt-readable histories, code, and rationales as conditioning signals; unlike many selector-style hyper-heuristics, they need not choose only from a predefined library of low-level heuristics.

In Genetic Programming (GP), Geometric Semantic Genetic Programming (GSGP)~\cite{moraglio2012gsgp,vanneschi2014semantic} already gives ``semantic'' a behavioural meaning: variation can be defined with respect to program output rather than only tree syntax. Here the object being varied may still be a program, but the distinctive move is different: GSGP uses algebraic variation with formal structure in output space, whereas an LLM operator samples from an implicit prompt-conditioned distribution without algebraic closure or a fixed semantic metric.\footnote{Section~\ref{sec:lens} separates this behavioural use of ``semantic'' from the prompt-channel use developed here.}

Bayesian optimization~\cite{shahriari2016taking} is another nearby tradition for expensive black-box objectives: it learns an explicit surrogate model and acquisition rule over the objective landscape, whereas an LLM variation operator changes the proposal mechanism induced by the prompt. The distinction matters because BO uses learning to choose where to evaluate the objective next, whereas the LLM call is itself the variation operator that generates the next candidate artifact.

Neural combinatorial optimization~\cite{kool2019attention} is also related but structurally distinct, training learned models to construct or guide solutions directly. Our lens applies when the search-relevant conditioning signal is exposed in the text prompt; when that signal is absorbed into model weights, the method falls outside the prompt-conditioning view. In each case, the boundary is the prompt interface: in related fields either the operator structure is fixed, learning is done without readable prompt conditioning, or semantics are defined over representations rather than over text supplied to the LLM at variation time.

These boundaries leave the central design question intact: once the proposal step is a prompt-conditioned LLM call, classification depends on which prompt-readable signal conditions the next candidate and which artifact, if any, survives the call. The next section turns these two attributes into the conditioning--persistence lens.

\section{The organizing lens: conditioning and persistence}
\label{sec:lens}
Section~\ref{sec:background} has identified the prompt representation $r$ as the design variable in an LLM proposal distribution. This section turns this variable into the paper's organizing lens. We first provide rules for determining the dominant conditioning channel in the variation prompt of an existing method, namely, the prompt signal that most strongly shapes the next candidate artifact. Moreover, we add a persistence descriptor for what survives after the LLM call. Together, conditioning and persistence form the descriptive framework used throughout the rest of the paper. 

\begin{figure}[!t]
\centering
\begin{tikzpicture}[font=\footnotesize,>=stealth,
  box/.style={draw=black!30, rounded corners, align=left, inner sep=4pt, font=\footnotesize, fill=orange!15, text width=4.7cm}]
 \node[box,fill=blue!6] (bN) at (0,0) {{\ttfamily h1: 427\\h2: 411\\h3: 419}\\[2pt]{\itshape\color{black!60} score-only trajectory over heuristics}};
 \node[font=\bfseries\footnotesize,text=black!75,below=1pt of bN] {\texttt{Numeric}};
 \node[box] (bS) at (-3.55,2.9) {{\ttfamily def step(T,U):\\\hspace*{1em}c = nearest(T[-1],U)\\\hspace*{1em}return T+[c]}\\[2pt]{\itshape\color{black!60} executable construction rule}};
 \node[font=\bfseries\footnotesize,text=black!75,above=1pt of bS] {\texttt{Symbolic}};
 \node[box] (bL) at (3.55,2.9) {{\ttfamily ``NN assigns tours to clusters;\\try a savings merge;\\then apply 2-opt repair''}\\[2pt]{\itshape\color{black!60} strategy-level change to the heuristic}};
 \node[font=\bfseries\footnotesize,text=black!75,above=1pt of bL] {\texttt{Linguistic}};
 \draw[->,dashed,black!45,line width=0.9pt] (bN.north) -- (bS.south);
 \draw[->,dashed,black!45,line width=0.9pt] (bN.north) -- (bL.south);
 \node[font=\footnotesize,text=black!55,fill=white,inner sep=1.5pt] at (0,1.34) {more explicit};
 \draw[<->,dashed,black!35,line width=0.9pt] (bS.east) -- (bL.west) node[midway,above=1pt,fill=white,inner sep=1.5pt,font=\footnotesize,text=black!55]{incomparable};
\end{tikzpicture}
\caption{The conditioning lens made concrete: one TSP operator (``propose the next construction heuristic'') conditioned on numeric, symbolic code, or linguistic content. These alternatives change the prompt representation $r$ supplied at the variation step and therefore shape the induced proposal distribution shown in Fig.~\ref{fig:proposal}.}
\label{fig:climb}
\end{figure}
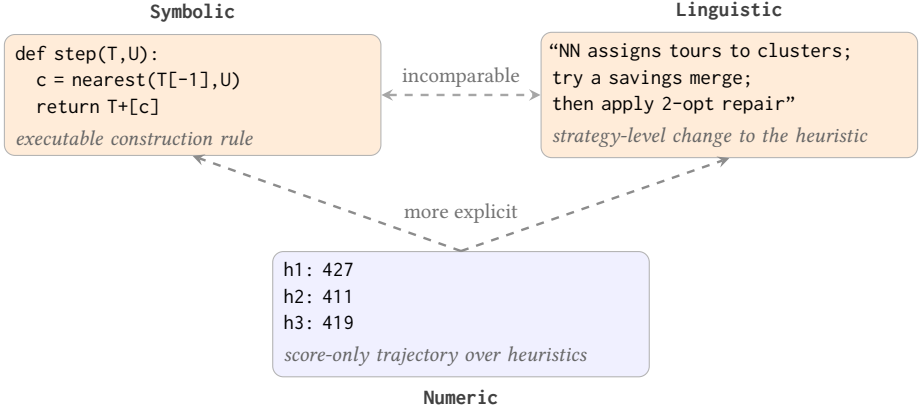

\paragraph{A dominant-conditioning typology.}
We classify methods by the dominant load-bearing channel in the prompt at the variation step (Fig.~\ref{fig:climb}). \texttt{Numeric} conditioning consists of the encoded search information familiar from classical metaheuristics: solution encodings, instance features, scalar scores, ranked parents, or score trajectories. \texttt{Symbolic} conditioning consists of machine-interpretable artifacts with formal structure---code, abstract syntax trees, formal rules, or structured graphs---whose syntax or behaviour can be checked by an external parser, validator, compiler, or executor. \texttt{Linguistic} conditioning consists of operative natural language (NL): dynamic critiques, reflections, diagnoses, design principles (which qualify even when reused across iterations, because their content encodes search-derived knowledge), or strategy notes that steer later proposals.\footnote{Fixed task headers and static problem statements do not by themselves make a method \texttt{Linguistic}.}

\paragraph{Hybrid methods and dominant channels.}
We regard the category \texttt{Numeric} intentionally in a broad way: a scalar score, a trajectory, and a feature vector all fall into \texttt{Numeric}, although they expose different amounts and kinds of search information. They share an evaluative role: they answer how candidate artifacts perform, although different summaries can affect behaviour differently without changing the conditioning class. \texttt{Numeric} is the base channel because it supplies the encoded state and objective feedback that metaheuristics already consume; \texttt{Symbolic} and \texttt{Linguistic} differ not only in surface form but in checkability: \texttt{Symbolic} content exposes behaviour that can be parsed, executed, or validated, whereas \texttt{Linguistic} content carries propositional or strategic rationale that can guide search but is not mechanically verified in the same way. Both are more explicit about problem structure than \texttt{Numeric}, but they are \emph{not} ranked against each other: code can be more structure-preserving than prose, while prose can express abstractions not present in the code. Most real methods are \textbf{hybrids} that use several channels at once. We therefore report the full set of channels present and classify a method by its \emph{dominant} one --- the channel that most drives the proposal. 

The analogy to black-box, gray-box, and white-box optimization is informative but not identical~\cite{santana2017graybox}: Santana's axis ranks how much of the \emph{problem's} structure is available, whereas our axis classifies the type of information placed in the LLM prompt. \emph{FunSearch} illustrates the distinction. The LLM receives prior program attempts selected by score, so the process is closer to gray-box than a pure scalar-only loop, yet on our axis it is \texttt{Symbolic} because the dominant conditioning signal is program code.

\paragraph{A reproducible rule for operator classification.}
A method's channel assignment is an operational label for design intent: it identifies which prompt signal is intended to drive variation. The rule consists of two steps. First, identify the prompt's operative, search-state-derived content. The \emph{operative-not-boilerplate} test filters out fixed headers and static problem statements: a header such as \texttt{[Task: Solve TSP]} does not lift a method to \texttt{Linguistic}, but a current reflection, diagnosis, or reusable principle can. Second, identify the channel intended to carry the operative search signal and audit that classification, where possible, by a \emph{drop-channel} ablation: remove or neutralize one channel at a time and ask which removal most changes the induced proposal while preserving a valid scaffold. The classification is therefore determined by the prompt signal that steers variation, not merely by the kind of artifact the method emits. When \texttt{Numeric} appears alongside operative \texttt{Symbolic} or \texttt{Linguistic} content, it is recorded as a supporting base channel. It becomes dominant only if that more explicit content fails the operative-not-boilerplate test. Likewise, a code artifact can be required for parsing or execution without being the dominant signal if the method's distinctive variation step is steered by an operative natural-language idea or reflection. If an ablation is inconclusive, report the source-described design-intent channel with the caveat; if two non-base channels both materially change proposals and neither clearly dominates, report a hybrid rather than forcing a single channel classification.

This ablation is an audit of the classification, not a clean causal attribution test. LLM outputs are sensitive to exact prompt phrasing, ordering, delimiters, decoding settings, and interactions among channels (Section~\ref{sec:background}). Single-channel ablations can also miss complementarities between channels, so the rule is a placement audit and information-signal diagnostic, not an estimate of independent causal contribution. The procedure therefore classifies the researcher's intended conditioning signal rather than inferring which tokens the model internally weights most. Assignments are reproducible for a given task and operator configuration; a different implementation could receive a different classification. Section~\ref{sec:tour} applies this rule in the method survey. Appendix~\ref{app:audit} adds a description-level agreement check: three LLM-based coders --- one instance each from three different model families --- classified the twelve representative methods of Table~\ref{tab:eval} from their channel descriptions, agreeing unanimously on nine of the twelve methods (Gwet's AC1~\cite{gwet2008ac1} $= 0.73$). The disagreements occur where channel dominance is least determinate, including a case already marked as a boundary case. This is a reliability check for the tutorial's representative classifications, not a claim that every method in the broader literature is equally unambiguous.

As a worked example, consider a ReEvo-style generator--reflector loop~\cite{ye2024reevo}. The variation prompt contains the current code, its score, and a natural-language reflection about why the code should change. The code is necessary: dropping it removes the executable artifact to be edited. But the distinctive operative signal is the reflection: neutralizing it reduces the method toward code-only automatic heuristic design, while preserving the scaffold and score feedback. The classification is therefore \texttt{Linguistic}-dominant with \texttt{Symbolic} support. This example demonstrates the classification procedure; it does not claim that the reflection channel is causally necessary for performance on every task.

\paragraph{Where the term ``semantic'' comes from.}
The word ``semantic'' follows the classical syntax/semantics contrast: syntax is surface form, while semantics is the meaning a representation carries~\cite{strachey2000fundamental,winskel1993formal}. Classical variation often manipulates the syntactic form of an encoded state through a hand-designed move, neighborhood, mutation, or construction rule; its domain knowledge is built into the operator's logic. Prompt-conditioned LLM variation changes that interface by exposing problem-relevant meaning in $r$, the prompt representation that conditions the next candidate. The three channels then name different kinds of such meaning. Numeric conditioning carries \emph{evaluative} meaning --- ``how good is this?'' \texttt{Symbolic} conditioning carries \emph{behavioural/denotational} meaning --- ``what does this compute or do?''\footnote{This is the sense closest to Geometric Semantic Genetic Programming (GSGP), where semantics is program behaviour rather than syntactic form~\cite{moraglio2012gsgp,vanneschi2014semantic}.} \texttt{Linguistic} conditioning carries \emph{propositional} meaning --- ``why does this work, and what should change?'' Therefore, the tutorial extends the syntax/semantics contrast from behavioural \texttt{Symbolic} content to evaluative \texttt{Numeric} and propositional \texttt{Linguistic} prompt content. It is not a claim that the model understands the problem, nor an NLP-style embedding-similarity claim.

\begin{figure}[!t]
\centering
\begin{tikzpicture}[font=\footnotesize,>=stealth,
  b/.style={draw=black!30,rounded corners,minimum height=7mm,minimum width=20mm,align=center,font=\footnotesize},
  llm/.style={b,fill=white},
  loop/.style={b,fill=blue!6},
  target/.style={b,draw=orange!60,dashed,fill=orange!10}]
 \node[font=\bfseries\footnotesize] at (1.2,2.88){\texttt{Transient}};
 \node[font=\footnotesize,text=black!60] at (1.2,2.58){call every iteration};
 \node[llm] (t1) at (1.2,1.8){\textbf{LLM}\\[-1pt]{\scriptsize [in-loop]}};
 \node[loop] (t2) at (1.2,0.0){\textbf{Search Loop}};
 \draw[->,black!50,line width=0.9pt] ([xshift=-4pt]t1.south) -- node[midway,left=3pt,font=\footnotesize,text=black]{artifact} ([xshift=-4pt]t2.north);
 \draw[->,black!50,line width=0.9pt] ([xshift=4pt]t2.north)  -- node[midway,right=3pt,font=\footnotesize,text=black]{state / score} ([xshift=4pt]t1.south);

 \node[font=\bfseries\footnotesize] at (5.4,2.88){\texttt{Amortized}};
 \node[font=\footnotesize,text=black!60] at (5.4,2.58){$\sim$0 runtime cost};
 \node[llm] (a1) at (5.4,1.8){\textbf{LLM}\\[-1pt]{\scriptsize [offline]}};
 \node[loop] (a2) at (5.4,0.0){\textbf{Search Loop}};
 \draw[->,black!50,line width=0.9pt] (a1)-- node[midway,right=3pt,font=\footnotesize,text=black]{artifact} (a2);
 \draw[->,black!50,line width=0.9pt] ([yshift=4.5pt]a2.west)
   .. controls +(-16pt,7pt) and +(-16pt,-7pt) ..
   node[pos=0.5,left=4pt,font=\footnotesize,text=black]{iterates} ([yshift=-4.5pt]a2.west);

 \node[font=\bfseries\footnotesize] at (10.25,2.88){\texttt{Transfer} (frontier)};
 \node[font=\footnotesize,text=black!60] at (10.25,2.58){re-bind / reuse};
 \node[llm] (f0) at (9.0,1.8){\textbf{LLM}\\[-1pt]{\scriptsize [distil]}};
 \node[loop] (f2) at (9.0,0.0){\textbf{Domain A}};
 \node[target] (f3) at (11.5,0.0){\textbf{Domain B}};
 \draw[<->,black!50,line width=0.9pt] (f0.south)--(f2.north);
 \draw[->,dashed,orange!70!black,line width=0.9pt] ([xshift=5pt]f2.north) to[out=35,in=145,looseness=1.05] (f3.north);
 \node[font=\footnotesize,text=black,align=center,fill=white,inner sep=1pt] at (10.35,1.08){artifact / principle};

 \coordinate (persistTtop) at (1.2,-0.58);
 \coordinate (persistTlow) at (1.2,-0.96);
 \coordinate (persistAinTop) at (5.25,-0.58);
 \coordinate (persistAinLow) at (5.25,-0.96);
 \coordinate (persistAoutTop) at (5.55,-0.58);
 \coordinate (persistAoutLow) at (5.55,-0.96);
 \coordinate (persistXtop) at (10.25,-0.58);
 \coordinate (persistXlow) at (10.25,-0.96);
 \draw[->,dashed,black!35,line width=0.9pt,shorten <=2.5pt,shorten >=3pt]
   (persistTtop) -- (persistTlow) -- (persistAinLow) -- (persistAinTop);
 \node[font=\footnotesize,text=black!55,above=2pt] at ($(persistTlow)!0.5!(persistAinLow)$){artifact survives};
 \draw[->,dashed,black!35,line width=0.9pt,shorten <=2.5pt,shorten >=3pt]
   (persistAoutTop) -- (persistAoutLow) -- (persistXlow) -- (persistXtop);
 \node[font=\footnotesize,text=black!55,above=2pt] at ($(persistAoutLow)!0.5!(persistXlow)$){artifact travels};
\end{tikzpicture}
\caption{
What persists, as a descriptive attribute. \texttt{Transient}: the LLM remains inside the loop. \texttt{Amortized}: an offline call emits an artifact that later runs at near-zero runtime cost. \texttt{Transfer}: a source-domain artifact or principle is reused in a new domain without rediscovering it.}
\label{fig:persist}
\end{figure}
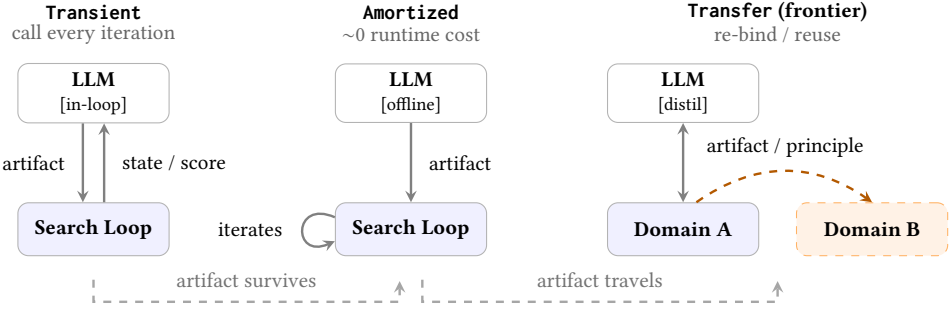

\paragraph{A second descriptor: persistence.}
Persistence --- what comes out and survives the call --- is conceptually independent of conditioning, though the two are empirically correlated in current methods (Fig.~\ref{fig:persist}). A \texttt{Transient} artifact is used within a single loop iteration; the loop therefore pays inference latency and token/API cost whenever it asks the model for a new candidate. An \texttt{Amortized} artifact is emitted offline and then runs at near-zero runtime inference cost, as a reusable operator, heuristic, or program. The frontier case is \texttt{Transfer}: an inspectable artifact or principle distilled from source-domain evidence is reused in a target domain without repeating the discovery process. \texttt{Transfer} is not new in spirit---classical cross-domain hyper-heuristics moved selection strategies across domains without an LLM~\cite{burke2013hyper}; the LLM-specific question is whether generated artifacts or principles are inspectable and re-bindable enough to travel robustly.

Persistence therefore bundles two practical subproperties: the \emph{call-locus} (is the LLM in the loop or called offline?) and the \emph{portability} of the surviving artifact. They usually align, but not always: executable artifacts may be re-bound directly, whereas prompt-mediated principles may still require model calls when applied to a new domain. In executable transfer, the artifact itself can run after its interface is adapted to the target evaluator. In principle transfer, the surviving object is a verbal rule or design rationale that seeds a new prompt and must then be validated on the target task.

Conditioning and persistence together characterize the methods surveyed in Sections~\ref{sec:tour} and~\ref{sec:choose}. They are separate descriptors but often coupled in current methods: solution-level outputs tend to be \texttt{Transient}, operator or program outputs tend to be \texttt{Amortized}, and principle-like outputs are natural candidates for \texttt{Transfer}. These are tendencies, not constraints. Section~\ref{sec:build} now instantiates the descriptors in a runnable prompt-conditioned operator.

\section{Developing LLM variation operators}
\label{sec:build}
This section turns the descriptors of Section~\ref{sec:lens} into a runnable prompt-conditioned operator. The construction has four components: prompt assembly, sampling, parsing and validation with bounded repair, and loop integration. By the end of the section, the reader has a runnable scaffold; the full copyable prompt template and a Python reference loop are in Appendix~\ref{app:templates}.

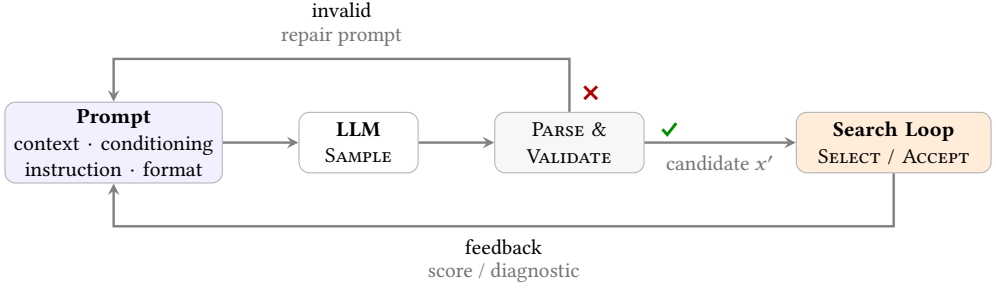
\begin{figure}[!t]
\centering
\begin{tikzpicture}[font=\footnotesize,>=stealth,
  b/.style={draw=black!30,rounded corners,minimum height=8mm,align=center,inner sep=2.5pt,font=\footnotesize},
  prompt/.style={b,fill=blue!6,text width=27mm},
  llm/.style={b,fill=white,text width=14mm},
  check/.style={b,fill=black!3,text width=18mm},
  loop/.style={b,fill=orange!15,text width=24mm}]
 \node[prompt] (p) {\textbf{Prompt}\\ context $\cdot$ conditioning\\ instruction $\cdot$ format};
 \node[llm,right=10mm of p] (l) {\textbf{LLM}\\ \textsc{Sample}};
 \node[check,right=10mm of l] (v) {\textsc{Parse} \&\\ \textsc{Validate}};
 \node[loop,right=20mm of v] (s) {\textbf{Search Loop}\\ \textsc{Select} / \textsc{Accept}};
 \draw[->,black!50,line width=0.9pt] (p)--(l);
 \draw[->,black!50,line width=0.9pt] (l)--(v);
 \draw[->,black!50,line width=0.9pt] (v)--
   node[pos=0.5,below=1pt,font=\footnotesize,text=black!55]{candidate $x'$}(s);
 \coordinate (validmark) at ($(v.east)!0.16!(s.west)+(0,5pt)$);
 \draw[green!55!black,line width=1.1pt]
   ($(validmark)+(-2.4pt,0.0pt)$) -- ($(validmark)+(-0.6pt,-2.0pt)$) --
   ($(validmark)+(2.8pt,2.3pt)$);
 \coordinate (repairv) at ([yshift=7mm]v.north);
 \coordinate (repairp) at (p.north |- repairv);
 \coordinate (invalidmark) at ($(v.north)!0.36!(repairv)+(8pt,0)$);
 \draw[red!70!black,line width=1.1pt]
   ($(invalidmark)+(-2.4pt,-2.4pt)$) -- ($(invalidmark)+(2.4pt,2.4pt)$)
   ($(invalidmark)+(-2.4pt,2.4pt)$) -- ($(invalidmark)+(2.4pt,-2.4pt)$);
 \draw[->,black!50,line width=0.9pt] (v.north) -- (repairv) --
   node[midway,above=1pt,font=\footnotesize,align=center]{\textcolor{black}{invalid}\\[-1pt]\textcolor{black!55}{repair prompt}}
   (repairp) -- (p.north);
 \coordinate (feeds) at ([yshift=-7mm]s.south);
 \coordinate (feedp) at (p.south |- feeds);
 \draw[->,black!50,line width=0.9pt] (s.south) -- (feeds) --
   node[midway,below=1pt,font=\footnotesize,align=center]{\textcolor{black}{feedback}\\[-1pt]\textcolor{black!55}{score / diagnostic}}
   (feedp) -- (p.south);
\end{tikzpicture}
\caption{Anatomy of an LLM variation operator: prompt, sample, parse-and-validate (with a repair path for invalid output), and loop integration with an optional feedback channel.}
\label{fig:anatomy}
\end{figure}

\paragraph{Anatomy.}
An LLM variation operator has four parts (Fig.~\ref{fig:anatomy}): a \emph{prompt} (problem context $\cdot$ conditioning content $\cdot$ instruction $\cdot$ output format); the \emph{sample} call; a \emph{parse-and-validate} stage that checks schema correctness and feasibility, returning a diagnostic on failure; and the \emph{loop integration} that feeds accepted candidates to selection and, optionally, passes feedback (a score or a diagnostic note) into the next prompt. Algorithm~\ref{alg:llmop} formalizes this loop.

\begin{algorithm}[!t]
\caption{LLM variation operator with bounded repair}
\label{alg:llmop}
\footnotesize
\begin{algorithmic}[1]
\REQUIRE $x_0$, $f$, $R$, $C$, step budget $T$, repair cap $K \ge 0$, stopping rule \textsc{Stop}
\STATE $x_{\mathrm{cur}} \gets x_0$; $s_{\mathrm{cur}} \gets f(x_{\mathrm{cur}})$
\STATE $x_{\mathrm{best}} \gets x_{\mathrm{cur}}$; $s_{\mathrm{best}} \gets s_{\mathrm{cur}}$
\STATE $H \gets [\,]$; $t \gets 0$
\WHILE{$t < T$ and not \textsc{Stop}$(x_{\mathrm{cur}},x_{\mathrm{best}},H,t)$}
  \STATE $r \gets \textsc{AssemblePrompt}(R,C,x_{\mathrm{cur}},s_{\mathrm{cur}},H)$
  \FOR{$k=0,\ldots,K$}
    \STATE $\mathrm{out} \gets \textsc{SampleLLM}(r)$
    \STATE $(ok,y,err) \gets \textsc{Validate}(\mathrm{out},R,C)$
    \IF{$ok$} \STATE \textbf{break} \ENDIF
    \STATE $r \gets \textsc{RepairPrompt}(r,\mathrm{out},err)$
  \ENDFOR
  \IF{not $ok$}
    \STATE $H \gets H \mathbin{\Vert} [(\mathrm{invalid},err)]$
    \STATE $t \gets t+1$
    \STATE \textbf{continue}
  \ENDIF
  \STATE $s_y \gets f(y)$
  \STATE $\Delta_y \gets s_y - s_{\mathrm{cur}}$
  \IF{\textsc{Accept}$(y,s_y,x_{\mathrm{cur}},s_{\mathrm{cur}},t)$}
    \STATE $x_{\mathrm{cur}} \gets y$; $s_{\mathrm{cur}} \gets s_y$
    \STATE $H \gets H \mathbin{\Vert} [(\mathrm{accepted},s_y,\Delta_y)]$
    \IF{$s_{\mathrm{cur}}$ better than $s_{\mathrm{best}}$}
      \STATE $x_{\mathrm{best}} \gets x_{\mathrm{cur}}$; $s_{\mathrm{best}} \gets s_{\mathrm{cur}}$
    \ENDIF
  \ELSE
    \STATE $H \gets H \mathbin{\Vert} [(\mathrm{rejected},s_y,\Delta_y)]$
  \ENDIF
  \STATE $t \gets t+1$
\ENDWHILE
\RETURN $x_{\mathrm{best}}$
\end{algorithmic}
\end{algorithm}

Algorithm~\ref{alg:llmop} formalizes the reference loop. The inputs (see \textbf{Require}; line~1) are the initial artifact~$x_0$, the objective function~$f$, the representation~$R$ and feasibility constraints~$C$ that together define a valid candidate, a step budget~$T$, a repair cap~$K$, and a stopping rule \textsc{Stop}. Here $T$ is the maximum number of main-loop steps, not the number of accepted candidates or repair attempts: each step assembles one prompt, may spend up to $K{+}1$ LLM samples because of bounded repair, and counts as one step against $T$ whether or not a candidate is accepted. The lowercase $r$ is the assembled prompt representation used in $q(x'\mid r)$; the uppercase $R$ is the candidate representation or schema used by the validator.

The first statements initialize the incumbent~$x_{\mathrm{cur}}$, the best-so far~$x_{\mathrm{best}}$, their cached scores, the history list~$H$, and the step counter~$t$. The history~$H$ stores validator diagnostics and selection outcomes; in the template below, it fills the ``Scores and diagnostics'' slot. The outer loop runs until the step budget is spent or \textsc{Stop} is triggered. The stopping rule represents the usual metaheuristic termination condition, such as a time limit, target quality, convergence test, or user interruption; any dependence on the initial artifact, wall-clock time, or a strict call cap can be captured inside \textsc{Stop}. Here $t$ counts main-loop steps, while $k$ indexes repair attempts within the current step.

Each step first assembles the prompt representation~$r$ from the current state: the representation, the constraints, the incumbent and its cached score, and the history gathered so far. The bounded-repair inner loop then samples a raw completion~$\mathrm{out}$ from the LLM, and \textsc{Validate} parses and checks the output, returning a valid artifact~$y$ only on success. On success the operator leaves the inner loop. On failure, \textsc{RepairPrompt} resends the current prompt together with the failing output and the validator's diagnostic before re-sampling, for at most $K{+}1$ attempts. Because each failed attempt can add text to the repair prompt, implementations should cap or summarize the repair block when context grows (see Section~\ref{sec:crosscut}). If the inner loop exhausts its retries without a valid candidate, the step is logged as invalid and the validator diagnostic is appended to $H$. The step is then skipped: it consumes one budget step but does not accept a candidate.

For a valid candidate $y$, the objective computes $s_y$, and $\Delta_y$ records its score gap to the incumbent. The selection rule may depend on the step counter~$t$ for any schedule or rule that changes over time; greedy acceptance can ignore that argument. An accept replaces the incumbent and its cached score, records the accepted candidate's score and gap, and updates the best-so-far from the accepted incumbent when $s_{\mathrm{cur}}$ is better than $s_{\mathrm{best}}$. A reject records the candidate score and gap, so the next prompt can use both successful and failed attempts as feedback. This feedback provides a mechanism for adaptation across steps rather than treating each proposal as independent of prior search outcomes. In either case, the outer budget counter is increased. On termination, the loop returns the best artifact found.

\paragraph{Output schema and validator.}
The \textsc{Validate} call inside the bounded-repair loop is itself a structured procedure; Algorithm~\ref{alg:validate} specifies it before we instantiate the prompt template. The validator turns the output contract into executable checks and returns a compact diagnostic when a check fails. Here $\mathrm{name}(R)$ is the expected value of the \texttt{representation} field for the chosen representation~$R$, e.g.\
\texttt{permutation} or \texttt{python}.
\begin{algorithm}[H]
\caption{\textsc{Validate}$(\mathrm{out},R,C)$}
\label{alg:validate}
\footnotesize
\begin{algorithmic}[1]
\STATE $m \gets \textsc{ParseEnvelope}(\mathrm{out})$
\IF{$m$ is incomplete} \RETURN $(false,\bot,$ ``schema error'') \ENDIF
\IF{$m.\mathrm{representation}$ does not match $\mathrm{name}(R)$}
  \RETURN $(false,\bot,$ ``representation mismatch'') 
\ENDIF
\STATE $y \gets \textsc{ParsePayload}(m.\mathrm{payload},R)$
\IF{$y$ is parse error} \RETURN $(false,\bot,$ ``syntax error'') \ENDIF
\IF{not \textsc{Feasible}$(y,C)$ within timeout}
  \RETURN $(false,\bot,$ \textsc{ExplainViolations}$(y,C))$ 
\ENDIF
\RETURN $(true,y,\emptyset)$
\end{algorithmic}
\end{algorithm}
Algorithm~\ref{alg:validate} makes the three layers explicit as a sequence of guarded checks. The procedure stops at the first failed layer and returns a diagnostic specific to that layer. The envelope and representation checks run first: \textsc{ParseEnvelope} extracts the structured candidate record from the raw completion using the envelope format declared by the prompt template; an incomplete record produces a \emph{schema error}, and an answer labeled \texttt{python}, for example, is rejected before a permutation parser tries to read it. Payload parsing follows: \textsc{ParsePayload} converts the payload --- the raw artifact text inside the envelope --- into a typed artifact~$y$ according to $R$, e.g.\ a permutation, an assignment, or a Python function, while malformed payloads return a \emph{syntax error}. Feasibility is checked last: an infeasible candidate returns concrete violations, e.g.\ a duplicated city or a capacity violation, which is far more useful to a repair prompt than a generic ``invalid.'' Only a candidate that clears all three layers is returned as valid. A timeout is treated as a feasibility failure; its limit should be set relative to the evaluator budget so that one malformed candidate cannot consume the run. The ordering is deliberate: cheap structural checks run before expensive feasibility checks, and because each layer emits its own diagnostic, the \textsc{RepairPrompt} call in Algorithm~\ref{alg:llmop} can target exactly the failure that occurred. For code artifacts, \textsc{ParsePayload} can perform syntactic checks such as abstract-syntax-tree (AST) parsing and import allow-listing. These checks reject malformed code and obvious undeclared dependencies, but they are not a security sandbox and do not rule out unsafe behavior through allowed names, dynamic execution patterns such as \texttt{eval} or \texttt{exec}, unsafe deserialization, or allowed modules that expose file, network, or system access. Execution-time safety belongs in \textsc{Feasible}: code should be type-checked and run in a separate process or container with restricted builtins, blocked file/network access, and time and memory limits. For direct solutions, the same \textsc{Feasible} layer applies the domain-specific constraint checker defined by $C$.

\paragraph{A reusable prompt template.}
With the loop and validator specified, the template defines what the designer fills in for each instantiation. It instantiates the prompt block of Fig.~\ref{fig:anatomy}, while Algorithms~\ref{alg:llmop} and~\ref{alg:validate} specify the sampling, validation, repair, and loop-integration blocks. The prompt has four sections and can be instantiated for solution proposals, code mutations, and heuristic design; the section roles stay fixed, while the representation~$R$, constraints~$C$, context, and feedback are filled with domain-specific content.
\begin{lstlisting}[style=prompt]
[context]
Problem type:
Objective direction: minimize | maximize
Candidate artifact: solution | operator | program
Available primitives: <functions, encodings, libraries, or moves>

[conditioning]
Representation R: <syntax of a candidate>
Constraints C: <hard feasibility rules>
Incumbent or parents: <artifact(s)>
Scores and diagnostics: <local evaluator output, failures, diversity notes>
Structural notes: <problem decomposition, known bottlenecks>

[instruction]
Emit exactly one new candidate artifact. Keep the representation R.
Use only the listed primitives where possible.
If diagnostics are present, target them.

[format]
Return only:
CANDIDATE
id: <short_id>
artifact_type: solution | operator | program
representation: <declared_representation>
payload:
<candidate in representation R>
END_CANDIDATE
\end{lstlisting}
Here the \texttt{CANDIDATE}/\texttt{END\_CANDIDATE} markers are the concrete envelope convention that \textsc{ParseEnvelope} expects; an implementation using JSON or structured output would replace only this format convention and its parser. The \texttt{payload} field contains the candidate itself. The following two traces instantiate this template in different ways: the first instantiates $R$ as a direct TSP permutation, so the payload is a solution (a tour); the second instantiates $R$ as Python code for a bin-packing heuristic, so the payload is a program.

\begin{figure}[!htbp]
\centering
\begin{tikzpicture}[font=\footnotesize,>=stealth,
  bub/.style={draw=black!30,rounded corners,inner sep=5pt,font=\footnotesize,align=left,text width=5.0cm},
  role/.style={draw=orange!70!black,rounded corners=1.5pt,fill=white,minimum width=14mm,
    minimum height=9mm,align=center,font=\footnotesize\bfseries,line width=0.8pt}]
 \begin{scope}[shift={(0,1.72)},draw=blue!55!black,line width=0.8pt]
   \fill[white] (0,0.2) circle (2.6mm); \draw (0,0.2) circle (2.6mm);
   \fill[white] (-0.50,-0.6) arc (180:0:0.50 and 0.48) -- cycle;
   \draw (-0.50,-0.6) arc (180:0:0.50 and 0.48);
 \end{scope}
 \node[font=\footnotesize,text=black!60] at (0,0.82) {designer};
 \node[bub,fill=blue!6,anchor=west] (p) at (1.15,1.5)
   {\textbf{Improve this TSP tour}\\
    Incumbent \texttt{[0,2,3,4,1]}\\
    length $9.236$\\
    keep each city exactly once};
 \node[role] (opbox) at (0,-0.78) {LLM};
 \draw[orange!70!black,line width=0.7pt]
   (-0.7,-0.66)--(-0.86,-0.66) (-0.7,-0.90)--(-0.86,-0.90)
   (0.7,-0.66)--(0.86,-0.66)  (0.7,-0.90)--(0.86,-0.90);
 \node[font=\footnotesize,text=black!60] at (0,-1.45) {operator $q(x'\mid r)$};
 \node[bub,fill=orange!15,anchor=west] (r) at (1.15,-0.95)
   {\textbf{Candidate}\\
    \texttt{[0,1,2,3,4]}\\
    length $8.000$\\
    valid, shorter, \emph{accepted}};
 \draw[->,black!50,line width=0.9pt] (p.south) to[out=-90,in=90,looseness=0.85]
   node[right=1pt,font=\footnotesize,text=black]{sample $\cdot$ validate} (r.north);
\end{tikzpicture}
\caption{An illustrative human--operator exchange: the \emph{designer} authors the prompt, and the \emph{LLM operator} samples a candidate that passes validation and improves the objective value.}
\label{fig:exchange}
\end{figure}
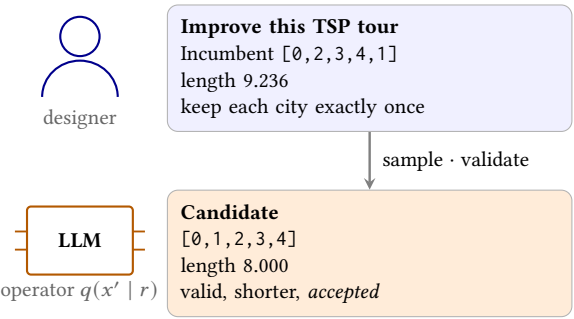

\paragraph{One traced iteration.}
The following is a simple illustrative build trace. It instantiates the framework as a \texttt{Numeric}-conditioned, \texttt{Transient} operator: the prompt contains a tour and score, the LLM proposes a successor solution, and the artifact is consumed inside the loop. Consider a five-node Euclidean TSP with coordinates $0=(0,0)$, $1=(1,0)$, $2=(2,0)$, $3=(2,2)$, $4=(0,2)$, incumbent tour \texttt{[0,2,3,4,1]}, and evaluated tour length $9.236$. Fig.~\ref{fig:exchange} summarizes the exchange. The listings show the prompt, sampled output, validation, repair, and acceptance steps for one outer iteration of Algorithm~\ref{alg:llmop}.

\begin{lstlisting}[style=prompt]
PROMPT SENT
[context] Illustrative TSP instance; minimize Euclidean tour length.
[conditioning] R = permutation of [0,1,2,3,4] starting at 0.
Node coordinates: 0:(0,0) 1:(1,0) 2:(2,0) 3:(2,2) 4:(0,2).
C = every city exactly once; closed tour length is evaluator score.
Incumbent: [0,2,3,4,1], score 9.236.
[instruction] Emit one lower-length tour if possible.
[format] Use the CANDIDATE envelope.

SAMPLED OUTPUT 
CANDIDATE
id: t1
artifact_type: solution
representation: permutation
payload:
[0,1,4,4,2]
END_CANDIDATE

VALIDATE
schema: ok; representation: permutation (match); payload: ok
feasibility: fail; city 4 duplicated and city 3 missing
\end{lstlisting}
Reading the first listing top to bottom: \textsc{AssemblePrompt} in Algorithm~\ref{alg:llmop} produces the \textsc{Prompt Sent} block --- the four sections of the template filled in for this instance. \textsc{SampleLLM} returns the \textsc{Sampled Output}: a well-formed \texttt{CANDIDATE} envelope whose payload is \texttt{[0,1,4,4,2]}. \textsc{Validate} then runs Algorithm~\ref{alg:validate}: the envelope and declared representation pass, and the list parses, but the feasibility layer finds city~4 duplicated and city~3 missing, so it returns the diagnostic ``city 4 duplicated and city 3 missing.'' This is a \emph{schema-valid but infeasible} candidate: $ok$ is false, the inner loop does not break, and \textsc{RepairPrompt} resends the full original context and appends the failing payload plus diagnostic:
\begin{lstlisting}[style=prompt]
REPAIR PROMPT (original context prepended; repair block shown)
The previous payload:
[0,1,4,4,2]
failed validation: city 4 duplicated and city 3 missing.
Emit one corrected CANDIDATE envelope. Do not change the representation.

REPAIRED OUTPUT
CANDIDATE
id: t1_fix
artifact_type: solution
representation: permutation
payload:
[0,1,2,3,4]
END_CANDIDATE

VALIDATE
schema: ok; representation: permutation (match); payload: ok
feasibility: ok
\end{lstlisting}
The second attempt ($k{=}1$) re-enters the inner loop: \textsc{SampleLLM} now returns the repaired output with payload \texttt{[0,1,2,3,4]} in the \texttt{REPAIRED OUTPUT} block. The \texttt{VALIDATE} block shows that the envelope, representation, payload, and feasibility checks now pass, so Algorithm~\ref{alg:validate} returns $(true,y,\emptyset)$ and the inner loop breaks. The objective scores the candidate at length $8.000$ on this illustrative instance; since $8.000 < 9.236$, the acceptance rule returns true: the incumbent $x_{\mathrm{cur}}$ becomes the repaired tour, the best-so-far $x_{\mathrm{best}}$ is updated from that incumbent, and \texttt{(accepted, 8.000, -1.236)} is appended to the history~$H$, so the next prompt is conditioned on the improvement. Had the repaired score been no better than $9.236$, the candidate would still be valid but rejected by selection, and the history would record the same gap under a rejected label. One outer step of Algorithm~\ref{alg:llmop} thus consumes, in this trace, two LLM samples and one accepted candidate.

\paragraph{A second trace: a code-emitting operator.}
The same loop can also instantiate a structurally different operator: a \texttt{Symbolic}, \texttt{Amortized} one that emits a reusable heuristic rather than a solution (Section~\ref{sec:lens}). For online bin packing (bin capacity~10), the operator emits a Python \texttt{priority(item, bins)} that ranks the open bins, and the evaluator scores a heuristic by the total number of bins it uses over a fixed set of instances with a lower bound of~19 bins. Starting from a first-fit heuristic (22 bins, $15.8\%$ above the bound), one outer iteration conditions the prompt on that parent program and its score:
\begin{lstlisting}[style=prompt]
PROMPT SENT
[context]
Problem type: online bin packing
Bin capacity: 10
Candidate artifact: program
Available primitives: arithmetic, comparisons, enumerate, list comprehension, float("-inf")

[conditioning]
Representation R: python function priority(item, bins)
Constraints C: return one numeric priority per open bin; no imports
Parent program:
def priority(item, bins):
    return [1.0/(i+1) if b >= item else float("-inf")
            for i, b in enumerate(bins)]
Score from external evaluator: 22 bins; lower bound 19
Diagnostic from external evaluator: first-fit leaves avoidable residual capacity

[instruction]
Emit one replacement priority function that prefers feasible bins likely
to reduce the total number of bins. Preserve the signature exactly.

[format]
Return only:
CANDIDATE
id: <id>
artifact_type: program
representation: python
payload:
<python function>
END_CANDIDATE
\end{lstlisting}
The LLM then samples a candidate program and validates it --- here \textsc{ParsePayload} parses the payload into an AST and rejects imports, and \textsc{Feasible} runs it in a restricted namespace (Algorithm~\ref{alg:validate}) --- before the evaluator scores it:
\begin{lstlisting}[style=prompt]
SAMPLED OUTPUT (LLM)
CANDIDATE
id: h1
artifact_type: program
representation: python
payload:
def priority(item, bins):
    # best-fit: prefer the tightest valid bin
    # infeasible bins receive -inf priority
    return [-(b - item) if b >= item else float("-inf") for b in bins]
END_CANDIDATE

VALIDATE (external)
schema: ok; representation: python (match); payload (AST): no imports
feasibility: ok (returns one score per input bin)

EVALUATE (external)
total bins = 19  (gap 0.0%, lower bound reached)  ->  accepted
\end{lstlisting}
The emitted best-fit heuristic reaches the lower bound for this illustrative instance set (19 bins) and is kept; it then runs with \emph{no} further LLM calls --- the cost is \texttt{Amortized}, in contrast to the per-move cost of the TSP operator above. Both traces are illustrative examples constructed in advance; their execution can be reproduced with the companion code released with this tutorial.

\paragraph{Design choices.}
The template leaves several choices to the designer; these should be reported because they determine the conditioning channel and persistence descriptor of Section~\ref{sec:lens}.
\begin{itemize}
\item \textbf{In-loop vs.\ offline.} Calling the LLM every iteration gives a \texttt{Transient} operator: it is often simpler to implement and can adapt to the current search state, but incurs one model call per move. Emitting an artifact offline gives an \texttt{Amortized} operator: it front-loads the model cost, and is preferable when a reusable artifact can be evaluated cheaply many times and then deployed without runtime model calls.
\item \textbf{Model choice.} The model is part of the operator design. Open-weight models are easier to pin, self-host, and run on private instances; API or private models may provide stronger completions but add version drift, data-exposure concerns, and per-token cost. Smaller models are generally cheaper and faster, whereas larger models often improve one-shot language-model quality and may reduce malformed completions in general settings~\cite{kaplan2020scaling,hoffmann2022chinchilla,abdin2024phi3}; for an operator, the trade-off should be measured on the target schema. The choice also couples to persistence: transient operators pay per call, while offline-amortized operators can spend a stronger model during design and deploy the emitted artifact without model access.
\item \textbf{Feedback.} No feedback gives near-independent samples; score feedback gives \texttt{Numeric} conditioning; diagnostics or reflections can add \texttt{Linguistic} conditioning. Use richer feedback when diagnostics are reliable, but cap or filter it when noisy messages could steer the model toward spurious fixes or exhaust the context window.
\item \textbf{Prompt layout.} Keep operative conditioning content close to the \texttt{[instruction]} slot rather than buried in a long \texttt{[context]} block; if the history grows, summarize or repeat the load-bearing signal near the end of the prompt (Section~\ref{sec:crosscut}).
\item \textbf{Output level.} Direct solutions are easy to parse and validate but usually transient. Programs and operators are more reusable and often \texttt{Symbolic}, but require a parser or compiler, sandboxed execution, unit tests, and resource limits.
\item \textbf{Selection.} Classical acceptance keeps the evaluator auditable and cheap. An LLM-based evaluator adds model calls and can introduce model bias; use it only when objective evaluation is unavailable or intentionally part of the task.
\end{itemize}

The same scaffold also exposes recurring failure modes: invalid outputs, diversity collapse, token and repair growth, context overflow, prompt sensitivity, and benchmark exposure. These are not separate operator families but cross-cutting risks of prompt-conditioned variation. Section~\ref{sec:crosscut} collects the corresponding mitigation and reporting discipline, while Section~\ref{sec:open} treats contamination and benchmark design as open infrastructure problems.

\section{Classification of representative methods according to the lens}
\label{sec:tour}
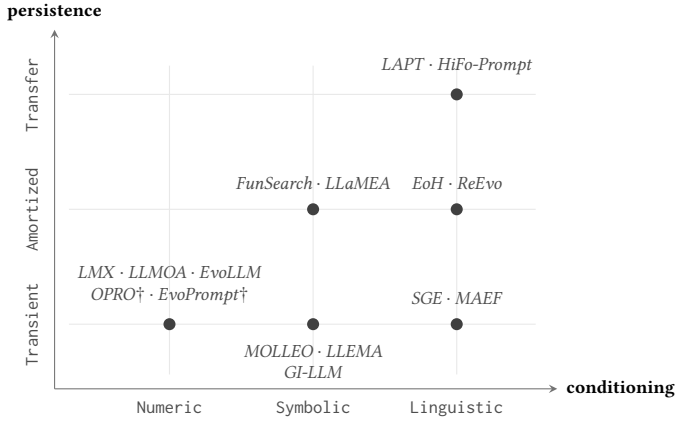
\begin{figure}[!t]
\centering
\begin{tikzpicture}[x=0.95cm,y=0.95cm,font=\scriptsize,>=stealth,
  dot/.style={circle,fill=black!75,inner sep=1.6pt},
  ml/.style={font=\scriptsize,text=black!70}]
 \draw[->,black!55] (0,0) -- (7.0,0) node[right,font=\scriptsize,black]{\textbf{conditioning}};
 \draw[->,black!55] (0,0) -- (0,5.0) node[above,font=\scriptsize,black]{\textbf{persistence}};
 \foreach \x in {1.6,3.6,5.6} \draw[black!9] (\x,0.2)--(\x,4.5);
 \foreach \y in {0.9,2.5,4.1} \draw[black!9] (0.2,\y)--(6.7,\y);
 \foreach \x/\l in {1.6/\texttt{Numeric}, 3.6/\texttt{Symbolic}, 5.6/\texttt{Linguistic}}
   \node[below=1pt,font=\scriptsize,text=black!70] at (\x,0){\l};
 \foreach \y/\l in {0.9/\texttt{Transient}, 2.5/\texttt{Amortized}, 4.1/\texttt{Transfer}}
   \node[rotate=90,font=\scriptsize,anchor=south,text=black!70] at (-0.1,\y){\l};

 \node[dot] at (1.6,0.9){};
 \node[dot] at (3.6,0.9){};
 \node[dot] at (5.6,0.9){};
 \node[dot] at (3.6,2.5){};
 \node[dot] at (5.6,2.5){};
 \node[dot] at (5.6,4.1){};

 \node[ml,anchor=south,align=center] at (1.6,1.05)
   {\emph{LMX}\,$\cdot$\,\emph{LLMOA}\,$\cdot$\,\emph{EvoLLM}\\
    \emph{OPRO}\(\dagger\)\,$\cdot$\,\emph{EvoPrompt}\(\dagger\)};
 \node[ml,anchor=north,align=center] at (3.6,0.75)
   {\emph{MOLLEO}\,$\cdot$\,\emph{LLEMA}\\\emph{GI-LLM}};
 \node[ml,anchor=south,align=center] at (5.6,1.05)
   {\emph{SGE}\,$\cdot$\,\emph{MAEF}};
 \node[ml,anchor=south] at (3.6,2.65)
   {\emph{FunSearch}\,$\cdot$\,\emph{LLaMEA}};
 \node[ml,anchor=south] at (5.6,2.65)
   {\emph{EoH}\,$\cdot$\,\emph{ReEvo}};
 \node[ml,anchor=south] at (5.6,4.25)
   {\emph{LAPT}\,$\cdot$\,\emph{HiFo-Prompt}};
\end{tikzpicture}
\caption{Representative method classifications on the conditioning--persistence map (the lens) of Section~\ref{sec:lens}. The horizontal axis records what conditions the LLM call; the vertical axis records what survives it. The \(\dagger\) symbol marks prompt-optimization boundary cases whose placement depends on prompt design.}
\label{fig:placement}
\end{figure}

After demonstrating how different LLM-assisted operators can be built, we now classify representative published methods through the lens outlined in Section~\ref{sec:lens}. Following the spirit of reference surveys in the field of metaheuristics~\cite{blum2003metaheuristics}, we classify methods, give representative ones a common skeleton, and place them in a shared frame. We first consider \texttt{Numeric}, \texttt{Symbolic}, and \texttt{Linguistic} operators, then discuss boundary and hybrid cases and map representative methods in Fig.~\ref{fig:placement}.

\paragraph{How we select and incorporate methods.}
Our selection is \emph{pedagogical, not systematic}. We do not aim to provide a complete census of the literature; recent systematic reviews, including a PRISMA review of this area, already serve that purpose~\cite{daros2026llm4co}. Instead, we include a method in case at least one of the following aspects is fulfilled: (i)~it can be classified and mapped to a \emph{distinct point on the conditioning lens}, so the channels are populated by real instances; (ii)~it introduces a \emph{distinct mechanism} worth learning about (island population, reflection, $(1{+}1)$ mutation, tree search, multi-objective selection); or (iii)~it serves as a \emph{background anchor} a newcomer needs, such as ELM, an open-ended-evolution precursor treated here as context rather than as a recommendation for scalar-objective optimization~\cite{lehman2022elm}.

Every method we discuss is read against its \emph{primary source}, so the skeletons (Algorithm~\ref{alg:ahd}) and the classifications (Fig.~\ref{fig:placement}) are faithful to those sources; in those cases in which various methods from the literature can be mapped to the same place on the organizing lens, we deal with representatives only and point to the systematic review for the full census. The result is a selective, verifiable map---not exhaustive by design. We \emph{exclude}, conversely: (a)~methods that are nearly equivalent to chosen representatives; (b)~systems that do not \emph{vary} a candidate inside a loop --- e.g.\ natural-language-to-model translators or one-shot optimization-modeling assistants, which are adjacent but out of scope; and (c)~any method that could not be read against a primary source.

\paragraph{Applying the classification rule.}
For each method class below, we apply the classification rule from Section~\ref{sec:lens}: examine the operative prompt content, identify the dominant channel, and audit the label by drop-channel ablation; where ablation is inconclusive, we follow the source-described design intent. We state what each method class is conditioned on, what it emits, and its essential mechanism; full one-line classifications can be found in Appendix~\ref{app:audit}, compact pseudocode in Algorithm~\ref{alg:ahd}, method class positions on the lens in Fig.~\ref{fig:placement}, and a broader coverage map in Appendix~\ref{app:coverage}. The comparison in Section~\ref{sec:choose} makes the trade-offs between different method classes explicit. Among the representative methods in Table~\ref{tab:eval}, this yields three \texttt{Numeric}-dominant, four \texttt{Symbolic}-dominant, and five \texttt{Linguistic}-dominant methods; the broader coverage map shows the same rapid, recent expansion, with most methods appearing during 2023--2026.

\paragraph{Discrete vs.\ continuous problems.}
One structural distinction cuts across the method classes: the type of the considered optimization problem shapes \emph{where} the operator can act. In \emph{discrete} problems the candidate is already a structured combinatorial representation, so the operator can receive that representation and emit another solution or a heuristic directly, and feasibility is a structural check (``each city once''); most automatic heuristic design work lives here (TSP, bin packing, scheduling). In \emph{continuous} problems---where solutions are real-valued vectors---LLMs propose precise numbers poorly at scale and have no native gradient. The usual design response is to place the LLM at the algorithm-design level: it emits an \emph{algorithm} or operator, and a classical numerical routine handles the fine-grained search over vectors. Accordingly, continuous work --- such LLaMEA applied to the BBOB black-box suite~\cite{vanstein2025llamea} and FunBO's discovered acquisition functions for Bayesian optimization~\cite{aglietti2024funbo}---is dominated by symbolic-conditioned, amortized operators: the LLM emits the search machinery offline; a classical method runs it. The lens organizes this split: discrete problems support operators at every conditioning level, including direct solution edits, whereas continuous problems tend to shift the operator up to the symbolic (algorithm-design) level.

\begin{algorithm}[t]
\caption{Skeletons of three code-emitting automatic heuristic design methods.}
\label{alg:ahd}
\footnotesize
\vspace{5pt}
\textit{LLaMEA} --- $(1{+}1)$ algorithm mutation~\cite{vanstein2025llamea}
\begin{algorithmic}[1]
\STATE $\mathrm{best} \gets$ an initial algorithm
\WHILE{budget remains}
  \STATE $r \gets (\mathrm{best}, f(\mathrm{best}),$ validation error or performance feedback$)$
  \STATE $\mathrm{out} \gets \textsc{SampleLLM}(r)$ \COMMENT{mutates the current best algorithm}
  \STATE validate and score $\mathrm{out}$; \textbf{if} $f(\mathrm{out})$ is at least as good \textbf{then} $\mathrm{best} \gets \mathrm{out}$
\ENDWHILE
\RETURN $\mathrm{best}$
\end{algorithmic}
\vspace{6pt}
\textit{FunSearch} --- island program search~\cite{romeraparedes2024funsearch}
\begin{algorithmic}[1]
\STATE initialize islands of programs, each scored by an evaluator $f$
\WHILE{budget remains}
  \STATE pick an island; sample $r$ from its high-scoring programs (exemplars)
  \STATE $\mathrm{out} \gets \textsc{SampleLLM}(r)$ \COMMENT{proposes a new program conditioned on code exemplars}
  \STATE validate $\mathrm{out}$; \textbf{if} valid (parses, matches interface, executes) \textbf{then} insert $\mathrm{out}$ scored by $f$
  \STATE trim lowest-scoring programs when the island exceeds capacity
\ENDWHILE
\RETURN best program
\end{algorithmic}
\vspace{6pt}
\textit{EoH} --- co-evolved idea $+$ code~\cite{liu2024eoh}
\begin{algorithmic}[1]
\STATE initialize a population of (idea, code) pairs
\WHILE{budget remains}
  \STATE select parents; assemble $r$ from their (idea, code) pairs via explore or modify strategy
  \STATE $\mathrm{out} \gets \textsc{SampleLLM}(r)$ \COMMENT{emits a new natural-language idea paired with code}
  \STATE score $\mathrm{out}$ by $f$; retain population's best
\ENDWHILE
\RETURN best
\end{algorithmic}
\vspace{4pt}
\end{algorithm}

\subsection{\texttt{Numeric}-conditioned methods}
Most \texttt{Numeric}-conditioned methods in this survey are transient, solution-level operators that use LLM calls to vary encoded candidates inside the search loop; \emph{OPRO} and \emph{EvoPrompt} are text-artifact boundary cases discussed below. \emph{LMX} is closest to crossover: the prompt presents parent encodings, such as bit strings or sequences, and the LLM returns an offspring encoding~\cite{meyerson2024lmx}; \emph{LLMOA} is a nearby hyper-heuristic case in which Gemini acts as a high-level component that constructs optimization sequences over low-level heuristics~\cite{zhong2024llmoa}; and \emph{EvoLLM} keeps an evolutionary scaffold while using fitness-ranked solution vectors to steer the next update~\cite{lange2024evollm}. \emph{LMEA} is another in-loop evolutionary case: in its TSP demonstration, parent tours and tour lengths condition LLM crossover and mutation over candidate tours~\cite{liu2024lmea}. The strength of these methods lies in their breadth: the prompt does not require executable machinery or a verbal rationale. The trade-of is that constraints, locality, and representation-specific structure usually have to be enforced by the encoding, move set, or validator rather than expressed explicitly in the
prompt.

\emph{OPRO} and \emph{EvoPrompt} are interesting boundary cases because they optimize \emph{prompts} scored on reasoning benchmarks rather than solutions to classical optimization problems. They are included here because the load-bearing conditioning signal is numeric: scored candidates, trajectories, or parent fitness values drive the search, even though the artifact being emitted and varied is text rather than a numeric solution encoding. \emph{OPRO} prompts the LLM with previously generated candidates and their scores under a fixed task instruction~\cite{yang2024opro}. \emph{EvoPrompt} runs a genetic-algorithm/differential-evolution (GA/DE) skeleton in which the individuals are prompts, mutation/crossover are LLM calls, and scores guide parent selection~\cite{guo2024evoprompt}. A variant driven mainly by search-derived critiques or explanations would instead be \texttt{Linguistic}; a dagger (\(\dagger\) symbol, see Fig.~\ref{fig:placement}) marks this dependence on the prompt design.

\subsection{\texttt{Symbolic}-conditioned methods}
Where \texttt{Numeric} methods primarily condition proposals on evaluative signals, \texttt{Symbolic} methods condition them on formal artifacts that can be parsed, executed, or checked by an external tool. Most \texttt{Symbolic} methods in the current combinatorial optimization landscape are code-emitting automatic heuristic design systems.\footnote{Code-emitting does not by itself imply \texttt{Symbolic} dominance: \emph{EoH}, \emph{ReEvo}, and \emph{MEoH} emit code but are classified as \texttt{Linguistic}-dominant by the dominant-channel rule (Section~\ref{sec:lens}), because the decisive conditioning signal is a natural-language idea (\emph{EoH} and \emph{MEoH}) or reflection (\emph{ReEvo}) that steers code edits. We classify them as \texttt{Linguistic}-conditioned methods below.} Here the operator searches \emph{program space} rather than solution space: the LLM is conditioned on code and emits code---a heuristic, an operator, or a whole algorithm---that can be deployed and run without the LLM. Two preconditions define this method type. A \emph{fast automatic evaluator} must exist, because every candidate program is scored by executing it; and the emitted artifact must be executable and inspectable, which is what makes the cost \texttt{Amortized}---the LLM is paid once, offline, and the emitted program then runs at no LLM cost.

The members differ mainly in how they organize the search. \emph{FunSearch} evolves programs in island populations scored by an evaluator, maintaining high-scoring programs per island~\cite{romeraparedes2024funsearch}; \emph{AEL} is an early automatic-heuristic-design loop that evolves heuristic code with an LLM generator and an execution evaluator~\cite{liu2023ael}; \emph{LLaMEA} drives a $(1{+}1)$ evolutionary loop over algorithm code~\cite{vanstein2025llamea}; and \emph{LLM-LNS} evolves neighborhood-selection heuristics for large-neighborhood search in MILP while a second prompt-strategy layer revises how those heuristics are generated~\cite{ye2025llmlns}. Further systems apply the same offline code-improvement pattern to whole optimizers and to non-expert workflows~\cite{chaconsartori2025improving,chaconsartori2026coforall}. We classify LLM-LNS as \texttt{Amortized} because the evolved neighborhood-selection heuristic is the artifact reused by the LNS solver, even though the underlying LNS procedure still operates inside a solve loop.

A second group of \texttt{Symbolic} methods keeps the offline generate--evaluate--retain pattern but changes how proposals are organized: \emph{MCTS-AHD} organizes proposals through tree search~\cite{zheng2025mctsahd}, \emph{FunBO} searches over code that defines Bayesian-optimization acquisition functions~\cite{aglietti2024funbo}, and \emph{AlphaEvolve} uses an agentic code-search loop with automated evaluators and a program database to evolve larger algorithmic codebases~\cite{novikov2025alphaevolve}. Unlike single-function heuristic search, it proposes and tests edits across a codebase, stores evaluated programs, and uses automated feedback to filter subsequent proposals.\footnote{\emph{LLM4AD} provides a shared platform for several automatic-heuristic-design methods in this family~\cite{liu2024llm4ad}.}

\texttt{Symbolic} conditioning is not limited to program code. Non-code \texttt{Symbolic} artifacts appear when the LLM varies a formal representation that can be parsed and checked by external tooling. For example, \emph{MOLLEO} uses LLM mutation and crossover over SMILES molecular strings inside an evolutionary loop~\cite{wang2025molleo}, while \emph{LLEMA} proposes structured crystallographic specifications for multi-objective materials discovery, which are then checked and scored by materials-domain evaluators~\cite{abhyankar2026llema}. GI-LLM is the code-level counterpart: the LLM acts as a mutation operator inside genetic improvement, emitting source-code patches that are parsed and tested in the loop~\cite{brownlee2023gillm}. These cases also show that \texttt{Symbolic} conditioning need not be \texttt{Amortized}: when the formal artifact itself is the in-loop candidate, the persistence descriptor is \texttt{Transient}.

Algorithm~\ref{alg:ahd} is an abstraction of the shared generate--evaluate--retain loop. \emph{EoH} is included as a contrast case: although it emits code, the prompt signal steering each edit is \texttt{Linguistic}, so emitted artifact type alone does not determine the dominant conditioning channel.

\subsection{\texttt{Linguistic}-conditioned methods}
Here the operative signal is search-derived natural language (NL), typically a reflection on previous generations, a critique of a candidate, or an accumulated design principle. Because this text can appear at different levels of abstraction, three persistence settings are useful to distinguish: language-guided search (\texttt{Transient}), language-steered code generation (\texttt{Amortized}), and principle accumulation (\texttt{Transfer}). Language-guided search is the \texttt{Transient} form: \emph{Self-Guiding Exploration (SGE)} uses thought trajectories generated during solving to decompose, execute, and refine combinatorial problem instances, while \emph{MAEF} uses role-specialized agents whose reasoning and evaluator feedback steer schedule generation. For \emph{MAEF}, the role-agent reasoning is the load-bearing linguistic signal; the evaluator scores are supporting feedback rather than the dominant channel. Neither method emits a reusable artifact~\cite{iklassov2024sge,wang2025maef}. Language-steered code generation is usually \texttt{Amortized}: it emits executable code, but the edits are steered by NL ideas or critiques inside the offline design loop. \emph{EoH} co-evolves an idea with its code implementation; \emph{ReEvo} is the explicitly reflection-based case, inserting a reflection step between generator calls so short- and long-term critiques condition later code proposals; and \emph{MEoH} extends the idea--code pattern with multi-objective selection over heuristic candidates~\cite{liu2024eoh,ye2024reevo,yao2025meoh}. Portfolio and diversity extensions refine that pattern: \emph{EoH-S} emits a set of complementary heuristics, while \emph{HSEvo} adds harmony-search and genetic-algorithm diversity machinery to keep heuristic directions separated~\cite{liu2026eohs,dat2024hsevo}. The reflection inside an offline method such as \emph{ReEvo} is \texttt{Transient} within the meta-search, whereas the final emitted code is the \texttt{Amortized} artifact.

\emph{Principle accumulation} makes this rationale the persistent artifact. This is the route by which \texttt{Linguistic} methods most naturally approach \texttt{Transfer}: a verbalized rule can be inspected, edited, and re-bound to a new instance family or domain instead of rediscovered from scratch. Two representative \texttt{Transfer}-oriented examples are \emph{LAPT} and \emph{HiFo-Prompt}: \emph{LAPT} distills successful neural-architecture-search runs into verbalized design principles, while \emph{HiFo-Prompt} combines hindsight principles with a foresight module that monitors population dynamics such as stagnation and diversity collapse to choose exploration or exploitation prompts~\cite{zhou2025lapt,hifo2026}. Other \texttt{Transfer}-labeled examples include \emph{EvoPH}, which transfers co-evolved prompts and heuristic code; \emph{MTHS}, which studies hierarchical cross-task heuristic design; and \emph{EvolCAF}, which carries \emph{EoH}-style design ideas into cost-aware acquisition-function design~\cite{liu2025evoph,liu2026mths,yao2024evolcaf}. Their success depends on inspectability and re-bindability, and remains a claim to validate on the target task.

Across these settings, what NL adds over code is \emph{rationale}: it can record \emph{why} a change should help, preserve a design rule that can later be re-bound to a new instance family, or name a search direction that code and scores do not carry as directly. Such rationale can broaden the search over design ideas and help maintain diversity, but it does not make LLM-produced candidates correct by itself. The cost is verifiability: NL is the least checkable conditioning, so a wrong reflection can mislead the search, and the \emph{operative-not-boilerplate} test (see Section~\ref{sec:lens}) matters most here.

\subsection{Boundary and hybrids}
The cases described above assume that the LLM is the component proposing the next artifact. Several systems sit at the edge of that assumption. Memetic hybrids pair an LLM proposal with classical local search or repair~\cite{biomimetics2024memetic}; controller-style methods use the LLM to route or schedule operators, connecting to adaptive operator selection and hyper-heuristics~\cite{fialho2010aos,thierens2005pursuit,burke2013hyper,ropke2006alns}; and agentic solvers combine proposal, tool use, and orchestration during solving~\cite{heuragenix2025}.

For these cases, the classification rule of Section~\ref{sec:lens} remains unchanged: we classify the method by the prompt signal used to propose the varied artifact, and annotate controller decisions, local-search post-processing, or tool orchestration separately. This keeps the catalogue from treating every auxiliary signal as a new dominant channel.

\subsection{Classification overview}

Fig.~\ref{fig:placement} shows classifications (mappings) of representative methods mentioned above with respect to the conditioning and persistence descriptors of Section~\ref{sec:lens} (the lens). Read each marker as a coordinate pair: the horizontal position records the dominant content type of the prompt, while the vertical position records what survives the call. The conditioning axis carries no ranking: \texttt{Symbolic} and \texttt{Linguistic} expose different kinds of structure and are incomparable taken as a whole. The persistence axis records increasing generalization scope, but not method quality. In the specific dimension of representation-portability, however, the three content types suggest a qualitative gradient from evaluative through denotational to propositional content: evaluative signals are closely tied to the search state; denotational signals are tied to a representation and can often be verified by execution; propositional signals can travel furthest but are less mechanically checkable. No channel dominates the other.

The current literature nevertheless maps onto the lower-right part of the layout: the chosen representative methods are classified on or below the diagonal formed by \texttt{Numeric}/\texttt{Transient}, \texttt{Symbolic}/\texttt{Amortized}, and \texttt{Linguistic}/\texttt{Transfer}. The diagonal is a layout choice, not an ordering of the conditioning axis; rearranging the unordered columns would produce a different visual pattern without changing the underlying tendencies. The tendency has a structural explanation: the persistence axis measures increasing generalization scope---from one iteration, to an entire run within a domain, to new domains---and the conditioning channel constrains how far an artifact can generalize. Evaluative content is specific to the search state that produced it; denotational content is specific to the representation and evaluator it was built for; propositional content is the most re-bindable because a later LLM call can re-interpret a design principle in a new context. Higher persistence therefore tends to require more domain-independent conditioning---not as a law, but as a structural tendency visible in the current body of methods. The methods below the diagonal prevent that pattern from becoming a rule: \emph{MOLLEO}, \emph{LLEMA}, and \emph{GI-LLM} occupy the \texttt{Symbolic}/\texttt{Transient} position, while \emph{SGE} and \emph{MAEF} occupy \texttt{Linguistic}/\texttt{Transient} positions, so the diagonal describes where method families cluster, not where individual methods must fall.

The empty cells should therefore be read as gaps in the current research landscape, not impossibility claims. \texttt{Numeric}/\texttt{Amortized} and \texttt{Numeric}/\texttt{Transfer} methods have not yet been developed, but classical cross-domain hyper-heuristics show that score-based transfer is possible in non-LLM settings~\cite{burke2013hyper}. \texttt{Symbolic}/\texttt{Transfer} does not yet have a representative because denotational content typically names the objects and operations of the domain in which it was built; crossing domains can change those referents, requiring the artifact to be adapted rather than simply reused. Section~\ref{sec:open} treats such missing conditioning--persistence pairings as empirical research questions.

\section{Choosing an appropriate method: an evidence-based comparison}
\label{sec:choose}
After the classification exercise of the previous section, we now turn from description to choice: which family should a practitioner use under a particular evaluator, budget, and deployment constraint?

\begin{figure}[!t]
\centering
\resizebox{\linewidth}{!}{%
\begin{tikzpicture}[scale=0.76,transform shape,font=\footnotesize,>=stealth,
  gate/.style={draw=black!35,rounded corners=2pt,fill=yellow!14,align=center,
    text width=34mm,minimum height=8.5mm,inner sep=3pt,font=\footnotesize},
  stop/.style={draw=black!35,rounded corners=2pt,fill=black!5,align=center,
    text width=36mm,minimum height=8.5mm,inner sep=3pt,font=\footnotesize},
  head/.style={draw=black!35,rounded corners=2pt,align=center,font=\footnotesize\bfseries,
    text width=33mm,minimum height=8.5mm,inner sep=3pt},
  leaf/.style={draw=black!35,rounded corners=2pt,align=center,
    text width=32mm,minimum height=25mm,inner sep=2.5pt,font=\footnotesize},
  refine/.style={draw=black!35,dashed,rounded corners=2pt,align=center,
    text width=31mm,minimum height=18mm,inner sep=2.5pt,font=\scriptsize},
  lab/.style={font=\scriptsize,fill=white,inner sep=1pt,text=black,
    text height=1.5ex,text depth=.25ex},
  retrylab/.style={font=\scriptsize,fill=white,inner sep=1pt,text=black!60,
    text height=1.5ex,text depth=.25ex},
  rlab/.style={font=\scriptsize,fill=white,inner sep=1pt,text=black!60,align=center},
  mainarr/.style={->,black!50,line width=0.9pt},
  mainline/.style={black!50,line width=0.9pt},
  dasharr/.style={->,dashed,black!45,line width=0.9pt},
  dashline/.style={dashed,black!45,line width=0.9pt}]

 \node[gate] (need) at (0,0) {Does an LLM add\\a needed capability?};
 \draw[draw=black!35,fill=black!5,rounded corners=0.4pt]
   (-0.28,0.84) -- (0.28,0.84) -- (0,0.58) -- cycle;
 \node[stop] (classic) at (5.45,0) {Use classical /\\non-LLM design};

 \node[gate] (eval) at (0,-1.55) {Evaluation\\ pipeline ready?};
 \node[stop] (build) at (5.45,-1.55) {Build evaluation\\ infrastructure first};

 \node[gate] (speed) at (0,-3.10) {Many candidate\\ evaluations feasible?};
 \node[stop] (slow) at (5.45,-3.10) {Query-efficient or\\ surrogate-assisted\\ strategies};

 \node[gate] (cost) at (0,-4.65) {Runtime LLM\\ calls affordable?};

 \node[head,fill=orange!16] (off) at (-4.00,-6.27) {Offline artifact\\ {\scriptsize\normalfont\color{black!55} [\texttt{Amortized}]}};
 \node[head,fill=blue!10] (loop) at (4.00,-6.27) {In-loop operator\\ {\scriptsize\normalfont\color{black!55} [\texttt{Transient}]}};

 \node[gate,text width=39mm] (offq) at (-4.00,-7.77) {Need NL ideas/reflections\\ as explicit design objects?};
 \node[gate,text width=39mm] (loopq) at (4.00,-7.77) {Need the LLM to directly\\ modify candidates?};

 \node[leaf,fill=green!10] (code) at (-6.15,-10.47)
   {\begin{minipage}[t][20mm][t]{30mm}\centering
    \strut\textbf{Code-conditioned}\\[-2pt]{\color{black!50}\rule{25mm}{0.25pt}}\\[1.5pt]
    {\scriptsize \emph{FunSearch}, \emph{LLM-LNS}\\ \emph{LLaMEA}, \emph{FunBO}\\ \emph{MCTS-AHD}, \emph{AlphaEvolve}}
    \vfill{\scriptsize\normalfont\color{black!55} [\texttt{Symbolic}]}
    \end{minipage}};
 \node[leaf,fill=purple!10] (nl) at (-2.05,-10.47)
   {\begin{minipage}[t][20mm][t]{30mm}\centering
    \strut\textbf{Language-steered}\\[-2pt]{\color{black!50}\rule{25mm}{0.25pt}}\\[1.5pt]
    {\scriptsize \emph{EoH}, \emph{ReEvo}}
    \vfill{\scriptsize\normalfont\color{black!55} [\texttt{Linguistic}]}
    \end{minipage}};

 \node[leaf,fill=purple!10] (ctrl) at (1.95,-10.47)
   {\begin{minipage}[t][20mm][t]{30mm}\centering
    \strut\textbf{Language-guided}\\[-2pt]{\color{black!50}\rule{25mm}{0.25pt}}\\[1.5pt]
    {\scriptsize \emph{SGE}, \emph{MAEF}\\[-1pt] \emph{HeurAgenix}}
    \vfill{\scriptsize\normalfont\color{black!55} [\texttt{Linguistic}]}
    \end{minipage}};
 \node[gate,text width=40mm] (formq) at (6.25,-10.47) {Does the candidate have a\\ formal verifiable structure?};

 \node[leaf,fill=green!10] (formal) at (3.85,-13.22)
   {\begin{minipage}[t][20mm][t]{30mm}\centering
    \strut\textbf{Formal-artifact}\\[-2pt]{\color{black!50}\rule{25mm}{0.25pt}}\\[1.5pt]
    {\scriptsize \emph{LLEMA}, \emph{MOLLEO}\\ \emph{GI-LLM}\\ \textit{(domain-specific)}}
    \vfill{\scriptsize\normalfont\color{black!55} [\texttt{Symbolic}]}
    \end{minipage}};
 \node[leaf,fill=cyan!9] (sol) at (8.00,-13.22)
   {\begin{minipage}[t][20mm][t]{30mm}\centering
    \strut\textbf{Solution variation}\\[-2pt]{\color{black!50}\rule{25mm}{0.25pt}}\\[1.5pt]
    {\scriptsize \emph{LMX}, \emph{LMEA}\\ \emph{EvoLLM}\\ \emph{OPRO}\(\dagger\), \emph{EvoPrompt}\(\dagger\)}
    \vfill{\scriptsize\normalfont\color{black!55} [\texttt{Numeric}]}
    \end{minipage}};

 \node[refine,fill=purple!5] (transfer) at (-0.10,-13.82)
   {\begin{minipage}[t][15mm][t]{29mm}\centering
    \strut\textbf{Principle accumulation}\\[-2pt]{\color{black!50}\rule{23mm}{0.2pt}}\\[1.5pt]
    \emph{LAPT}, \emph{HiFo-Prompt}\vfill{\normalfont\color{black!55} [\texttt{Transfer}]}
    \end{minipage}};
 \node[refine,fill=purple!5] (diverse) at (-4.60,-13.82)
   {\begin{minipage}[t][15mm][t]{29mm}\centering
    \strut\textbf{Diversity-seeking}\\[-2pt]{\color{black!50}\rule{23mm}{0.2pt}}\\[1.5pt]
    \emph{EoH-S}, \emph{HSEvo}\\ \emph{MEoH}\vfill{\normalfont\color{black!55} }
    \end{minipage}};

 \coordinate (checky) at (0,-15.62);
 \coordinate (checktopl) at (code.west |- checky);
 \coordinate (checktopr) at (sol.east |- checky);
 \coordinate (checkcenter) at ($(checktopl)!0.5!(checktopr)$);
 \coordinate (checkbotr) at ([xshift=-1.30cm,yshift=-0.98cm]checktopr);
 \coordinate (checkbotl) at ([xshift=1.30cm,yshift=-0.98cm]checktopl);
 \draw[draw=black!35,fill=black!3,rounded corners=2pt]
   (checktopl) -- (checktopr) -- (checkbotr) -- (checkbotl) -- cycle;
\node[font=\footnotesize\bfseries,text=black] at ($(checkcenter)+(0,-0.31)$) {Verify};
 \node[font=\footnotesize,text=black!80] at ($(checkcenter)+(0,-0.66)$)
   {budget parity \(\cdot\) target-task validation \(\cdot\) drop-channel ablation};

 \draw[mainarr] (need.east) -- node[lab,above] {no} (classic.west);
 \draw[mainarr] (need.south) -- node[lab,right,pos=0.5] {yes} (eval.north);
 \draw[mainarr] ([yshift=6pt]eval.east) -- node[lab,above,pos=0.5] {no} ([yshift=6pt]build.west);
 \draw[mainarr] (eval.south) -- node[lab,right,pos=0.5] {yes} (speed.north);
 \draw[mainarr] ([yshift=6pt]speed.east) -- node[lab,above,pos=0.5] {no} ([yshift=6pt]slow.west);
 \draw[mainarr] (speed.south) -- node[lab,right] {yes} (cost.north);
 \draw[dasharr] ([yshift=-6pt]build.west) --
   node[retrylab,below,pos=0.50] {retry} ([yshift=-6pt]eval.east);
 \draw[dasharr] ([yshift=-6pt]slow.west) --
   node[retrylab,below,pos=0.50] {retry} ([yshift=-6pt]speed.east);

 \coordinate (costsplit) at ($(cost.south)!0.5!(cost.south |- off.north)$);
 \coordinate (offcostbend) at (off.north |- costsplit);
 \coordinate (loopcostbend) at (loop.north |- costsplit);
 \draw[mainline] (cost.south) -- (costsplit);
 \draw[mainarr] (costsplit) -- (offcostbend) -- (off.north);
 \draw[mainarr] (costsplit) -- (loopcostbend) -- (loop.north);
 \node[lab,above,yshift=1pt] at ($(costsplit)!0.65!(offcostbend)$) {no};
 \node[lab,above,yshift=1pt] at ($(costsplit)!0.65!(loopcostbend)$) {yes};

 \draw[mainarr] (off.south) -- (offq.north);
 \draw[mainarr] (loop.south) -- (loopq.north);

 \coordinate (offsplit) at ($(offq.south)!0.5!(offq.south |- code.north)$);
 \coordinate (codebend) at (code.north |- offsplit);
 \coordinate (nlbend) at (nl.north |- offsplit);
 \draw[mainline] (offq.south) -- (offsplit);
 \draw[mainarr] (offsplit) -- (codebend) -- (code.north);
 \draw[mainarr] (offsplit) -- (nlbend) -- (nl.north);
 \node[lab,above,yshift=1pt] at ($(offsplit)!0.64!(codebend)$) {no};
 \node[lab,above,yshift=1pt] at ($(offsplit)!0.64!(nlbend)$) {yes};

 \coordinate (loopsplit) at ($(loopq.south)!0.5!(loopq.south |- ctrl.north)$);
 \coordinate (ctrlbend) at (ctrl.north |- loopsplit);
 \coordinate (formqbend) at (formq.north |- loopsplit);
 \draw[mainline] (loopq.south) -- (loopsplit);
 \draw[mainarr] (loopsplit) -- (ctrlbend) -- (ctrl.north);
 \draw[mainarr] (loopsplit) -- (formqbend) -- (formq.north);
 \node[lab,above,yshift=1pt] at ($(loopsplit)!0.64!(ctrlbend)$) {no};
 \node[lab,above,yshift=1pt] at ($(loopsplit)!0.64!(formqbend)$) {yes};

 \coordinate (formsplit) at ($(formq.south)!0.5!(formq.south |- formal.north)$);
 \coordinate (formalbend) at (formal.north |- formsplit);
 \coordinate (solbend) at (sol.north |- formsplit);
 \draw[mainline] (formq.south) -- (formsplit);
 \draw[mainarr] (formsplit) -- (formalbend) -- (formal.north);
 \draw[mainarr] (formsplit) -- (solbend) -- (sol.north);
 \node[lab,above,yshift=1pt] at ($(formsplit)!0.64!(formalbend)$) {yes};
 \node[lab,above,yshift=1pt] at ($(formsplit)!0.64!(solbend)$) {no};

 \coordinate (divstart) at ([xshift=-6mm]nl.south);
 \coordinate (trstart) at ([xshift=6mm]nl.south);
 \draw[dashline] (divstart) -- ++(0,-0.92) coordinate (divstem);
 \draw[dashline] (trstart) -- ++(0,-0.92) coordinate (trstem);
 \coordinate (divbend) at (diverse.north |- divstem);
 \coordinate (trbend) at (transfer.north |- trstem);
 \draw[dashline] (divstem) --
   node[rlab,above,yshift=3pt,pos=0.50] {if diversity\\needed} (divbend);
 \draw[dashline] (trstem) --
   node[rlab,above,yshift=3pt,pos=0.50] {if transfer\\needed} (trbend);
 \draw[dasharr] (divbend) -- (diverse.north);
 \draw[dasharr] (trbend) -- (transfer.north);
\end{tikzpicture}
}
\caption{Decision guide for choosing whether and how to use an LLM variation operator. The verification bar summarizes the required reporting checks.}
\label{fig:decision-expanded}
\end{figure}
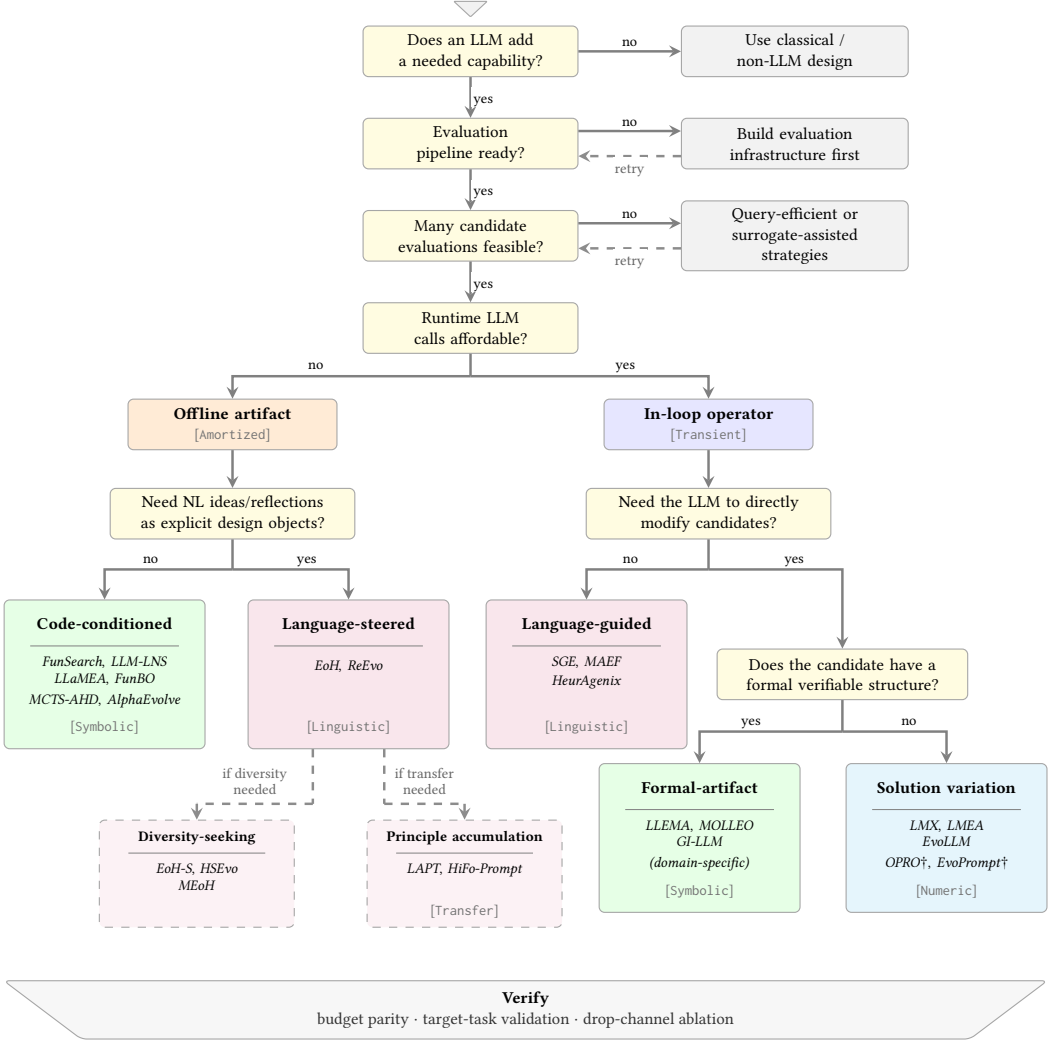

\paragraph{Relevant dimensions.}
Important factors when deciding on one of the outlined methodologies are the intended application, the available budget, and runtime setting. Five dimensions govern that choice, developed here through an evidence table (Table~\ref{tab:eval}) and a rule-based decision guide (Fig.~\ref{fig:decision-expanded}); based on the descriptors of Section~\ref{sec:lens} and the build choices of Section~\ref{sec:build}. The first is \emph{call-locus} (equivalently, cost-locus): whether the model is invoked during offline design or inside the search loop at each move or iteration. This choice matters because it changes where the model inference cost is paid and whether the deployed system remains model-dependent.\footnote{This is the closest analogue to algorithm selection~\cite{rice1976algselection,kotthoff2014algsel,kerschke2019algsel}: choose a method based on the features and constraints of the problem instance. Standard algorithm selection usually requires a shared performance matrix over methods, instances, and budgets; such a matrix does not yet exist for LLM variation operators.} The second is \emph{solution quality}, assessed here as source-specific evidence rather than a cross-method ranking: methods often serve different purposes, and even similar methods usually report on their own benchmarks and evaluators (Table~\ref{tab:eval}). The evidence therefore supports comparison against each source's own baselines more than comparison across rows~\cite{bartzbeielstein2020bench}. Because a shared performance matrix does not yet exist, the decision guide is rule-based: it asks practical questions about evaluator readiness, budget, call-locus, and persistence rather than ranking methods statistically. The third is \emph{LLM cost and latency}: this is the dimension classical metaheuristics usually do not have to account for. Invalid output and bounded repair can inflate the per-accepted-candidate cost substantially. Offline methods pay this cost during meta-search but may deploy a reusable artifact without model calls; in-loop methods pay during the run, often per iteration. Section~\ref{sec:crosscut} gives the full accounting list. The fourth is \emph{robustness and reproducibility}, which determine how much confidence to place in any single-run headline: stochastic decoding, prompt sensitivity, model/version drift, and evaluator implementation choices can all change the observed result. The fifth is \emph{artifact persistence}: whether the emitted artifact can be reused after generation without keeping the model available at serving time, or whether the optimization loop remains a permanent model-dependent system. The guide below routes mainly on evaluator readiness, cost, call-locus, and persistence; solution quality and reproducibility act as verification filters through the evidence table and the shared reporting controls.

\begin{sidewaystable}[ph!]
\rowcolors{2}{white}{gray!15}
\centering
\begin{adjustbox}{max totalsize={0.95\textheight}{0.92\textwidth},center}
\begin{minipage}{\textheight}
\caption{Evidence summary for representative LLM variation operators. Column \textbf{Conditioning} indicates the dominant conditioning channel. \emph{Call locus}: per op.\,=\,per operator application, per gen.\,=\,per generation, per step\,=\,per search step. \emph{Evidence}: PR\,=\,peer-reviewed, pre\,=\,preprint, own\,=\,source-specific benchmark, matched\,=\,same evaluator, qual\,=\,qualitative only. Rows marked \emph{pre} or \emph{own} require the corresponding caution when read outside their source setting. The \(\dagger\) symbol marks boundary cases.}
\label{tab:eval}
\small
\setlength{\tabcolsep}{4pt}
\renewcommand{\arraystretch}{1.14}
\begin{tabularx}{0.96\textheight}{@{}>{\raggedright\arraybackslash}p{2.15cm}
                  >{\raggedright\arraybackslash}p{1.9cm}
                  >{\raggedright\arraybackslash}p{2.05cm}
                  >{\raggedright\arraybackslash}p{1.75cm}
                  >{\raggedright\arraybackslash}p{1.55cm}
                  >{\raggedright\arraybackslash}X@{}}
\toprule
\textbf{Method} & \textbf{Conditioning} & \textbf{Emitted} & \textbf{Call locus} & \textbf{Evidence} & \textbf{Reported result} \\
\midrule
\emph{LMX} \cite{meyerson2024lmx} & \texttt{Numeric} &solution & per op. & PR; qual & demonstrates semantic crossover behavior on synthetic tasks; no headline metric on classical optimization benchmarks \\
\emph{OPRO} \cite{yang2024opro} & \texttt{Numeric}\(\dagger\) &prompt / solution & per step & PR; own & GSM8K up to $+8$ p.p.; some BBH tasks up to $+50$ p.p.\ vs.\ human prompts \\
\emph{EvoPrompt} \cite{guo2024evoprompt} & \texttt{Numeric}\(\dagger\) &prompt & per gen. & PR; own & some BBH tasks up to $+25$ p.p.\ vs.\ human \& auto prompts \\
\midrule
\emph{FunSearch} \cite{romeraparedes2024funsearch} & \texttt{Symbolic} &program & offline & PR; own & cap set $n{=}8$: $512$, bound $C\geq2.2202$; beats best-fit on online BPP \\
\emph{LLaMEA} \cite{vanstein2025llamea} & \texttt{Symbolic} &algo.\ code & offline & PR; own & BBOB subset at 5D: outperforms CMA-ES \& DE on several functions under reported budget \\
\emph{MCTS-AHD} \cite{zheng2025mctsahd} & \texttt{Symbolic} &program & offline & PR; matched & smallest matched online-BPP gap: $0.89\%$ vs.\ FunSearch/EoH/ReEvo \\
\emph{AlphaEvolve} \cite{novikov2025alphaevolve} & \texttt{Symbolic} &program / code & offline & pre; own & rank-48 tensor decomposition for $4{\times}4$ complex matmul; first characteristic-0 improvement over Strassen's rank-49 and 14 matrix targets improved \\
\midrule
\emph{EoH} \cite{liu2024eoh} & \texttt{Linguistic} &idea + code & offline & PR; own & online BPP: 20 generations / 2,000 LLM queries; matches or improves FunSearch in its own evaluation \\
\emph{ReEvo} \cite{ye2024reevo} & \texttt{Linguistic} &code & offline & PR; own & competitive on 6 CO problems, more sample-efficient \\
\emph{HSEvo} \cite{dat2024hsevo} & \texttt{Linguistic} &code population & offline & PR; matched & BPO/TSP/OP matched study: improves or matches FunSearch/EoH/ReEvo while tracking diversity \\
\emph{LAPT} \cite{zhou2025lapt} & \texttt{Linguistic} &principles & offline & PR; own & beats prior transferable-NAS (TNAS) methods on most tasks \\
\emph{MEoH} \cite{yao2025meoh} & \texttt{Linguistic} &pareto set of heuristics & offline & PR; own & up to $10\times$ more efficient; more trade-offs \\
\bottomrule
\end{tabularx}
\end{minipage}
\end{adjustbox}
\end{sidewaystable}

\paragraph{An evaluative comparison.}
With the dimensions identified, Table~\ref{tab:eval} evaluates representative methods on conditioning, emitted artifact, and call-locus, and reports each method's \emph{own} headline result. These entries are source-specific evidence, not a \emph{leaderboard}: the methods are evaluated on different benchmarks---extremal combinatorics, the BBOB black-box suite, online bin packing, and prompt-optimization tasks---and their headline numbers do not support direct cross-method ranking. The \emph{Evidence} column therefore marks whether a result is peer-reviewed or preprint evidence, and whether it is source-local, matched, or qualitative; source-specific and preprint rows are useful signals, but not settled comparative evidence. Matched evidence is available only in narrow AHD groups. In online bin packing, \emph{MCTS-AHD}, \emph{FunSearch}, \emph{EoH}, and \emph{ReEvo} are scored under the same evaluator: \emph{MCTS-AHD} achieves the smallest reported BPP gap ($0.89\%$), and \emph{EoH} reports a 20-generation online-BPP run (2,000 LLM queries) that matches or improves \emph{FunSearch} in the paper's own evaluation. \emph{HSEvo} provides a second matched comparison over BPO, TSP, and OP, emphasizing the diversity--quality trade-off. Even in these groups, the comparison is not a channel ablation: channel and search mechanism co-vary across the methods, and \emph{MCTS-AHD}'s advantage may reflect its tree-search structure rather than its \texttt{Symbolic} conditioning channel. The drop-channel ablation discipline of Section~\ref{sec:crosscut} is designed to partially separate such effects. The table is representative rather than exhaustive; some methods discussed in the survey are not assigned separate evidence rows.

\paragraph{Fair comparison requirements.}
A reported result should guide method choice only when the experimental controls support the particular comparison being made. At least, a credible within-study comparison should use the same evaluator and harness, compare against a strong classical baseline where available, include a no-LLM or drop-channel ablation, and match the relevant budgeted resources; Section~\ref{sec:crosscut} gives the full reporting list, including repair attempts and accepted candidates. Preprint and source-specific rows in Table~\ref{tab:eval} remain informative, but should be interpreted within their original experimental setting until comparable controls are available. Across benchmark sets, best-run claims should be replaced by repeated runs and appropriate statistical tests~\cite{bartzbeielstein2020bench,derrac2011nonparametric}. Section~\ref{sec:crosscut} discusses the failure modes these controls are intended to reveal.

\paragraph{A decision guide.}
The first branch is the null choice: add an LLM variation operator only when it adds a capability that a classical operator, hyper-heuristic, or algorithm-configuration method does not. The operative test is a capability gap, not model availability. Prefer a classical or non-LLM design when, for example, the move family for your local search is already well explored and tested, when inference or evaluation cost cannot be justified at the required search budget, or when model outputs cannot be automatically parsed, validated, and evaluated; in each case, the added LLM layer introduces cost and failure modes without a clear compensating role. An LLM earns its place when it can explore structured artifact spaces that are too large to enumerate manually, extract search guidance from heterogeneous signals---objective values, constraint violations, and natural-language annotations---that resist encoding as fixed operator logic, or define operator move families in domains where the appropriate rules are not known in advance. In all cases, measure the gain against a matched no-LLM baseline under the same evaluation budget; if the advantage is not retained under that control, additional search effort may explain the result. The no-LLM baseline therefore serves two roles: it is both an evaluation control and the first design gate.

Figure~\ref{fig:decision-expanded} turns this into a staged flowchart; Table~\ref{tab:eval} is its evidence companion. The \emph{Call locus} column maps to the locus gate, the \emph{Emitted} column names the deployable or runtime artifact, and the \emph{Evidence} and \emph{Reported result} columns state how source-specific each result is. The flowchart operationalizes call-locus and LLM cost as branching decisions; solution quality and reproducibility reappear in the verification step at the bottom. Three gate questions route the practitioner:
\begin{enumerate}
  \item Is an automated evaluation pipeline---including evaluator, validator, and repair mechanism---in place? This is a prerequisite for any iterative LLM-based search: without it, generated artifacts cannot be reliably checked, scored, or selected (Section~\ref{sec:build}, Algorithms~\ref{alg:llmop}--\ref{alg:validate}).
  \item Can many candidates be evaluated? This scale depends on the number of candidates considered per generation. Offline automatic heuristic design (in the style of \emph{FunSearch}, \emph{EoH}, \emph{LLaMEA}, or \emph{MCTS-AHD}) can place substantial pressure on evaluator throughput because meta-search repeatedly evaluates generated artifacts~\cite{romeraparedes2024funsearch,liu2024eoh,vanstein2025llamea,zheng2025mctsahd}. When throughput is limited, a query-efficient loop, careful evaluation budgeting, or surrogate/acquisition-based screening is required~\cite{aglietti2024funbo}.
  \item Is runtime LLM inference affordable within the wall-clock and token budget? Offline methods move model inference to a single design phase and can subsequently deploy a reusable artifact without further model calls, whereas in-loop methods incur model cost during optimization. 
\end{enumerate}
Once these questions have been answered, the remaining choices identify the operator family. On the offline path, the key question is whether natural-language ideas or reflections should be explicit design objects, or whether the population should contain only formal code artifacts. On the in-loop path, the key question is whether the LLM directly modifies candidate artifacts or supplies guidance that a conventional operator executes; if it modifies candidates directly, the candidate's formal verifiable structure further separates formal-artifact methods from numeric solution variation. These are structural distinctions: they follow from what the LLM emits and what the downstream pipeline consumes. The dashed refinements under Language-steered methods mark two special cases: principle accumulation is preferred when a domain-invariant principle can be identified and validated across problem types, whereas diversity-seeking variants are warranted when the proposal distribution collapses toward too-narrow a family of candidates (Section~\ref{sec:crosscut}). The binary branching is a readability simplification; in practice, hybrid schedules can call the model every \(k\) iterations, only at restarts, or only when the search stagnates. Within the selected family, final method choice still requires target-task validation, budget parity, and drop-channel ablation under the fair-comparison requirements above; Table~\ref{tab:eval} supplies source evidence but does not replace these checks. The verification bar in Fig.~\ref{fig:decision-expanded} summarizes this handoff from family selection to evidence validation.

The guide should therefore be read as a routing aid for family architecture, not as a universal selector or a complete experimental claim. It narrows whether the appropriate starting point is an offline artifact, an in-loop solution-variation operator, a language-guided controller, a transfer-oriented principle system, or a portfolio/diversity variant. The branches are binary for readability, but real deployments couple these choices with evaluator readiness, validation and repair, evaluation-call budget, token budget, wall-clock constraints, and reproducibility requirements. Section~\ref{sec:crosscut} addresses these shared evaluation and engineering risks, which must be resolved before any result drawn from the guide can be trusted.

\section{Cross-cutting concerns}
\label{sec:crosscut}
The choice of the LLM-based variation operator type is only the first step. The resulting method still inherits engineering and evaluation risks shared across the lens. The concerns below are operational rather than taxonomic: they affect \texttt{Numeric}, \texttt{Symbolic}, and \texttt{Linguistic} operators, and they should be addressed before any reported improvement is interpreted as evidence.

\paragraph{Prompt sensitivity and ablation discipline.}

The conditioning channel is not the whole prompt. Wording, example order, delimiters, schema fields, and the position of load-bearing content can alter LLM proposals even when the channel class is unchanged~\cite{zhao2021calibrate,lu2022fantastically,sclar2024formatting,mizrahi2024multiprompt}. Because models condition on the exact preceding text, not on an abstract representation of what that text means, long histories create a placement risk: decision-critical information may be used less reliably when buried in the middle of the context window~\cite{liu2023lostmiddle}. Critical conditioning should therefore remain close to the instruction, and any summarization or retrieval rule used to fit the context window should be treated as part of the prompt design and archived with the template. A credible study should release the exact template, test a small set of meaning-preserving variants, and distinguish formatting failures from search failures. Drop-channel ablation identifies the load-bearing signal, but it does not test sensitivity to wording, examples, or schemas within the same channel. When supported by the backend, structured-output APIs, tool/function calling, and grammar-constrained decoding can constrain generation and localize format failures~\cite{geng2023grammar,willard2023efficient}; validation remains necessary, and schema names or field descriptions may themselves carry instructional signals. Substantive schema changes should therefore be frozen or ablated alongside prompt variants. Hallucination is a related grounding failure: plausible content unsupported by the specification, API, codebase, or evaluator can introduce false constraints, invalid calls, spurious diagnoses, or unsafe code paths into the search state. Generated content should pass the relevant schema, validator, sandbox, source check, or local evaluation before conditioning later prompts. Lower temperature or narrower top-$p$ may reduce variance and format drift, but cannot prevent unsupported claims.

\paragraph{Diversity at the proposal step.}

Diversity collapse can occur at generation time: LLM priors from training, alignment, and decoding may make repeated calls produce similar candidates before acceptance or selection begins. This differs from evolutionary search, where diversity is typically reduced by selection, replacement, or local improvement; with an LLM operator, the proposed candidates may already be too similar before those mechanisms act. Downstream diversity mechanisms can therefore only act on the generated variants. Methods such as \emph{FunSearch}, \emph{EoH-S}, \emph{HSEvo}, and \emph{MEoH} address this through islands, set-valued output, diversity pressure, or multi-objective search~\cite{romeraparedes2024funsearch,liu2026eohs,dat2024hsevo,yao2025meoh}. Diversity can be measured by edit or syntactic distance, behavioral spread on probe instances, or objective-space coverage, and should be reported over time alongside objective value. Mitigation targets inputs (diverse examples or histories), populations (islands or distinct generators),\footnote{Distinct generators may use different prompts, model checkpoints, or even different models across islands.} and decoding (higher temperature or top-$p$ while preserving an acceptable valid-output rate~\cite{holtzman2020nucleus}).

\paragraph{Evaluation validity.}

Budget comparisons are often ambiguous: an ``evaluation'' may denote an objective call, an LLM sample, a token budget, a repaired/accepted candidate, or wall-clock time. These quantities are not directly comparable and should therefore be reported separately: evaluator calls, model calls/samples, input/output tokens, repair attempts, accepted candidates, and wall-clock.\footnote{\texttt{Transient} operators pay per use, amortized methods shift cost to meta-search, and batching or API latency can change runtime independently of the algorithm. For self-hosted inference, accelerator time and measured energy should be reported when available; for API models, calls and tokens are imperfect proxies for monetary and infrastructure cost~\cite{henderson2020energy,strubell2019energy}.} Budget parity is still insufficient if the scoring source changes: LLM judges can conflate search quality with evaluator preferences, especially when the generator and judge share a model family~\cite{zheng2023mtbench,shi2025judgingjudges}.\footnote{Position bias, style preference, and self-consistency effects can create circular evidence rather than objective improvement.} Thus, comparisons require identical evaluators, strong baselines, no-LLM or drop-channel controls, repeated runs, and appropriate statistical tests~\cite{bartzbeielstein2020bench,derrac2011nonparametric}. Contamination by pre-training data is a related community-level benchmarking issue, treated in Section~\ref{sec:open}; a practical guard is to run a zero-shot or held-out probe before treating search success on a public benchmark as evidence of generalization.

\paragraph{Latency, batching, and infrastructure.}

Embedding an LLM in the variation operator changes runtime shape: unlike local-search moves, LLM calls are remote, high-variance, and often latency- or quota-bound. Population methods can hide part of this latency through batching or asynchronous calls. Self-hosting shifts bottlenecks to local hardware and serving configuration, whereas API access adds rate limits, provider-side latency variance, prompt-logging/data-retention constraints, and possible moderation or safety filters; these issues matter especially for sensitive optimization data. Batching and asynchronicity help only when generated artifacts can pass automatically through parsing, feasibility checks, sandboxing, bounded repair, and diagnostics --- the infrastructure required by the first gate in Fig.~\ref{fig:decision-expanded}. For code-emitting or tool-using operators, execution safety belongs in the evaluation pipeline. In new domains, this infrastructure may be the main cost of entry.

\paragraph{Operator non-stationarity.}

LLM variation operators are time-dependent in a way most classical operators are not: model updates, endpoint changes, or decoding defaults can alter the induced proposal distribution, creating a moving-baseline problem~\cite{chen2023gptdrift}. Claims therefore depend on the model version, endpoint/checkpoint date, decoding settings, and repair policy used to instantiate the operator. This is especially important for \texttt{Transient} operators, which remain model-dependent at every run; thus, reproduction across model families or pinned checkpoints strengthens the evidence. For offline-\texttt{Amortized} methods, the model belongs to the design process rather than the deployed product. Releasing the generated heuristic, program, or operator, together with model, prompt, log, evaluator, seed, and environment details, is therefore central to reproducibility~\cite{gundersen2018repro,lopezibanez2021repro,pineau2021repro}. \\

Taken together, these concerns define the minimum requirements for credible LLM-in-meta\-heuristics studies: prompt and history handling, diversity reporting, matched evaluation controls, safe execution, and artifact release must be specified before a claimed improvement can be interpreted as operator evidence. Section~\ref{sec:open} asks what the field still lacks even after these conditions are met.

\section{Open problems and research directions}
\label{sec:open}
Section~\ref{sec:crosscut} identified recurring risks, mitigations, and reporting controls needed for credible LLM-in-metaheuristics studies. The directions below go beyond these baseline requirements: they remain open because the field still lacks the theoretical foundations, infrastructure, and empirical evidence needed to address them fully and reliably.

\paragraph{Theoretical foundations.}

The proposal-distribution notation $q(x'\mid r)$ makes design effects observable, but it is not yet a theory of LLM-induced search. We do not know which proposal properties --- support over feasible artifacts, locality, diversity, or probability mass near high-quality candidates --- would support finite-budget guarantees, nor how those properties depend on the conditioning channel in $r$. The natural object is the stochastic process induced by proposal, validation, repair, and acceptance --- a Markov chain when the search state is explicit. This separates asymptotic convergence from finite-budget performance, while allowing the effective proposal kernel to depend on prompt history, decoding, and validation or repair state. Classical theory gives useful anchors but also shows why broad claims would be premature: No-Free-Lunch results require assumptions on the problem class~\cite{wolpert1997nfl}, and convergence results for Simulated Annealing or Genetic Algorithms assume transition kernels, schedules, or variation mechanisms that LLM operators do not automatically satisfy~\cite{hajek1988cooling,rudolph1994convergence}. EDA and Bayesian views of in-context learning offer analogies, not guarantees~\cite{larranaga2002eda,xie2022iclbayes}. Similarly, GSGP provides a behavioral foundation for program variation without covering prompt-conditioned edits, natural-language rationales, or model priors~\cite{moraglio2012gsgp,vanneschi2014semantic}. The theory of LLMs and in-context learning is itself still partial, so a general optimization theory built on LLM-induced proposal kernels remains premature. Progress should therefore start with restricted claims: finite-budget results for validated LLM proposal families, or measurable conditions under which a channel increases valid-proposal mass or quality without collapsing diversity.

\paragraph{Benchmarking infrastructure.}

LLM-based metaheuristics need benchmark infrastructure tailored to prompt-conditioned operators, since classical test sets may not expose evaluator differences, unequal token or repair budgets, contamination through known instances or solver traces, and sensitivity to artifact-validation harnesses. Thus, a fair classical benchmark may still fail to isolate the LLM operator's contribution. Existing efforts provide partial components: contamination studies motivate optimization-specific audits~\cite{xu2024contamination}; \emph{LLM4AD} offers reusable generation/evaluation infrastructure~\cite{liu2024llm4ad}; \emph{COCO} exemplifies controlled optimization benchmarking~\cite{hansen2021coco}; and the online bin-packing results in Table~\ref{tab:eval} demonstrate same-evaluator comparison across LLM-based automatic-heuristic-design methods. The open problem is to integrate these practices across domains~\cite{bartzbeielstein2020bench}. A practical suite should include a parametric fresh-instance generator, shared evaluators, classical/no-LLM baselines, standardized accounting of evaluator/model calls, tokens, repairs, and time, multi-run statistical protocols~\cite{derrac2011nonparametric}, and contamination audits. Zero-/one-shot tests on hidden instances can reveal when search adds little beyond direct model capability, though they do not detect subtler benchmark familiarity. Robustness should also be tested under randomized, shifted, and adversarial instances, while transfer claims require cross-domain benchmarks with controlled source--target structural distance.

\paragraph{The cost-quality frontier.}

The question from Section~\ref{sec:intro} --- whether more explicit conditioning improves performance --- remains unresolved, with evidence suggesting task-dependent effects rather than a general rule. The key open direction is the cost-quality Pareto frontier: how solution or heuristic quality trades off against token budget, repair rate, valid-output rate, energy or accelerator usage, and deployment portability, with model scale as a study variable. A central hypothesis is that representation design and model scale can partly compensate for one another: schemas, executable structure, and sharper diagnostics can reduce model burden, whereas loose or verbose conditioning increases token and context demands.

Testing this requires matched factorial comparisons: fix the evaluator, outer loop, and total resources; vary conditioning channel, call-locus/persistence, and model scale; and report quality per token, valid-output rate, repair rate, diversity, latency, and energy or accelerator usage. Although model scale remains confounded with architecture, training data, alignment, and serving stack~\cite{kaplan2020scaling,hoffmann2022chinchilla}, matched comparisons can help separate representation effects from model-strength effects. \emph{EoH}, \emph{MEoH}, and \emph{LLaMEA} provide partial evidence on quality and model-call costs~\cite{liu2024eoh,yao2025meoh,vanstein2025llamea}. A first credible frontier study should test whether channel choice shifts the quality-per-token curve under matched budgets and at least two model scales.

\paragraph{Auditability and interpretability.}

Auditability asks whether conditioning signals actually shape search. The drop-channel rule of Section~\ref{sec:lens} provides channel-level attribution, while content-level ablations test which history segments matter within a channel. Removing failures, successes, or reflections while preserving the prompt format establishes counterfactual sensitivity~\cite{madsen2022posthoc}: it shows that a prompt component changes the output, not how the model computes that change. Artifact-level audits complement prompt ablations. Edit distance, AST similarity, and clustering can reveal recurring heuristic families, while linguistic principles can be tested behaviorally by comparing artifacts generated with and without them. Such analyses show whether prompt changes alter the structural or behavioral family of generated artifacts, not only the final score. Recent work shows that LLM representations of combinatorial problem structure can be recovered by probing~\cite{daros2025behavior}, but whether these representations actively shape $q(x'\mid r)$ during generation or are only recoverable after the fact remains unknown.

Interpretability asks the harder model-level question: how a large learned model turns the conditioning signal into a proposal. Chain-of-thought can expose stated reasoning~\cite{wei2022cot}, but it remains an output artifact rather than direct access to the internal computation that produced it. Circuit analysis and causal editing offer tools for identifying internal features and pathways~\cite{elhage2021circuits,meng2022rome}, yet their relevance to optimization artifacts remains open. Progress would require paired content ablations and artifact analyses showing that removing history components changes both performance and the structural or behavioral family of proposals. In agentic systems, attribution may instead target roles or tool calls.

\paragraph{Transfer and artifact portability.}

Transfer is the flagship persistence challenge: the goal is not merely more conditioning, but persisted artifacts that are inspectable and sufficiently instance- or domain-invariant to be re-bound without rediscovery. The sparse \texttt{Transfer} coverage in Fig.~\ref{fig:placement} and Table~\ref{tab:coverage} reflects that most systems demonstrate reuse within a task distribution rather than out-of-distribution transfer. Three forms are relevant. First, executable transfer carries symbolic artifacts such as heuristic code, which must be sufficiently decoupled from the source evaluator to adapt to a target domain; \emph{GEAKG} illustrates this direction~\cite{chaconsartori2026geakg}. Second, principle-level transfer carries declarative design rules through later prompts, as in \emph{LAPT}, \emph{HiFo-Prompt}, and \emph{EvoPH}; \emph{EvolCAF} similarly transfers an evolved acquisition-function design~\cite{zhou2025lapt,hifo2026,liu2025evoph,yao2024evolcaf}. Such principles may travel more easily than source-specific programs, but cannot be mechanically verified in the target domain. Third, meta-level transfer asks whether process-level knowledge for generating or selecting operators can bias or initialize search on new problem classes, echoing classical hyper-heuristics~\cite{burke2013hyper}; \emph{MTHS} provides an early LLM-based example~\cite{liu2026mths}. Progress would require controlled source--target tests showing that persisted artifacts reduce target-domain design effort relative to rediscovery, rather than merely demonstrating reuse after manual adaptation.

\paragraph{Agentic and multi-agent metaheuristics.}

Agentic systems turn a single LLM operator into a role-structured process involving generators, critics, verifiers, planners, tool users, or orchestrators. Multi-agent systems are the case where multiple communicating model instances or separately role-prompted calls interact~\cite{guo2024multiagent}. Among surveyed methods, \emph{ReEvo} uses a generator--reflector split~\cite{ye2024reevo}, while \emph{MAEF} and \emph{HeurAgenix} use richer role structures~\cite{wang2025maef,heuragenix2025}; \emph{AlphaEvolve} instead combines generation, a program database, and automated evaluation~\cite{novikov2025alphaevolve}.

Key open questions concern architecture and evaluation: which role compositions improve search, preserve diversity, and remain auditable at acceptable cost? Same-family critics may reproduce generator blind spots, embedding the LLM-as-judge circularity of Section~\ref{sec:crosscut} in the architecture. Role messages can propagate erroneous diagnoses, while diversity may collapse through generators, critics, or shared search vocabulary. Additional roles can multiply calls and repairs, and multi-role call graphs make failures harder to attribute~\cite{banerjee2025errattr}. Studies should therefore log call graphs, role-specific token costs, and accepted-artifact lineage. A convincing next step is to compare single- and multi-role systems under identical evaluators, budgets, and logging, isolating gains from role structure rather than extra calls, selection pressure, or blind-spot amplification.

\paragraph{Beyond prompt channels.}

Two frontiers extend beyond text-only prompt channels. Weight-internalized variation stores search signals in fine-tuned, adapter-based, or RL-trained parameters rather than inspectable prompts, as in \emph{EvoTune}~\cite{surina2025evotune}; related neural combinatorial optimization directly constructs or guides solutions~\cite{bengio2021ml,kool2019attention}. If conditioning resides only in weights, the drop-channel rule no longer applies, and domain-specific operators raise the same cross-domain transfer question: can they generalize without retraining?

Multimodal conditioning uses non-text representations of instances or evolving solutions: \emph{ViTSP} renders fixed problem instances, while \emph{VEO} renders evolving incumbents or solution states~\cite{yin2025vitsp,zhao2025veo}. The key question is whether vision adds information beyond an equivalent textual description under identical loops and budgets. Hybrid text--visual prompts further ask whether visual input contributes complementary structure or merely duplicates information. Progress requires controlled comparisons of textual, visual, and weight-internalized conditioning, with the visual or weight-internalized signal removed or replaced as a drop-channel analogue.

\section{Conclusion}
\label{sec:conclusion}
This tutorial organizes LLM variation operators in metaheuristics through a two-descriptor framework: dominant prompt conditioning and artifact persistence. Conditioning asks what the prompt makes load-bearing at the variation step---numeric evidence, symbolic structure, or linguistic rationale. Persistence asks what survives the model call---a transient candidate used inside the loop, an amortized artifact reused after offline design, or a transfer artifact re-bound to a new family or domain. In the classical syntax/semantics sense, this is the paper's ``semantic turn'': variation moves from classical operators---which apply fixed, hand-coded moves, mutations, construction rules, or recombination rules to encoded states---to prompt-conditioned LLM operators in which the representation \(r\) is supplied in the prompt at generation time and becomes an explicit design choice. The content supplied through \(r\) can therefore be characterized as evaluative (Numeric), behavioural or denotational (Symbolic), or propositional (Linguistic). This does not claim that the model understands the problem or the representations it manipulates, nor that richer or more verbose prompts necessarily produce better search behaviour.

Operationally, the framework gives the tutorial three uses: method placement, operator construction, and family selection. Placement is based on the prompt content that is load-bearing at generation time, not on the emitted artifact alone; the drop-channel audit makes that judgment reproducible. Construction treats the LLM call as one component in a complete variation loop: the prompt structure determines what information reaches the model; the completion format, parser, validator, and bounded-repair path determine which outputs become valid candidates; and the evaluator determines how accepted artifacts feed back into search. When choosing among alternatives, the null option remains the starting point: add an LLM only when it supplies a capability that classical operators, hyper-heuristics, or configuration methods cannot achieve. After that gate, the decision follows the practical constraints of the guide: whether the validation and evaluation infrastructure is ready, whether enough candidates can be evaluated, whether runtime calls fit the wall-clock and token budget, and whether the deployment setting favours offline artifact design or in-loop model use. Those answers determine the conditioning family the setting can support.

Reliable evidence requires more than correct operator placement. Prompt sensitivity, hallucinated or malformed outputs, diversity loss, evaluator mismatch, latency, infrastructure constraints, and model drift are risks that cut across all conditioning channels and method families. They call for a reporting discipline: generated artifacts should be validated, subject to bounded repair, and sandboxed when they execute code; comparisons should match evaluators, model budgets, and wall-clock limits; and studies should release prompts, artifacts, logs, and drop-channel or content-level audits. Several foundational gaps remain open: finite-budget theory for LLM-induced proposal distributions, contamination-resistant benchmark infrastructure, cost-quality evidence across conditioning richness and model scale, and transfer results that go beyond within-distribution reuse. Closing these gaps would make LLM-conditioned operators more than an expressive way to generate moves; it would make them a reproducible mechanism for extending metaheuristic search to problems where effective variation is difficult to hand-design.

\begin{acks}
Guillem Rodr\'iguez-Corominas was supported by the European Commission--\allowbreak{}NextGenerationEU through the Momentum CSIC Programme: Develop Your Digital Talent, under project MMT24-\allowbreak{}IIIA-02. 
\end{acks}

\bibliographystyle{ACM-Reference-Format}
\bibliography{references-seed_new}

\clearpage
\appendix
\section{Reusable templates}
\label{app:templates}
This appendix provides a copyable scaffold that instantiates the anatomy of Section~\ref{sec:build}
(Fig.~\ref{fig:anatomy}) and the build loop of Algorithm~\ref{alg:llmop}: first the prompt the operator
sends, then the code that wraps it.

\paragraph{The prompt template.}
The template consists of the four slots mentioned in Section~\ref{sec:build}; the angle-bracketed fields are the only
parts that change per problem. \textbf{[context]} (lines~1--5) fixes the task and the legal building
blocks --- line~4 declares what the operator emits (a solution, an operator, or a program), and line~5
bounds the output to a vocabulary the validator can check. \textbf{[conditioning]} (lines~7--12) is the
load-bearing channel (Section~\ref{sec:lens}): the representation $R$ and constraints $C$ (lines~8--9)
stay fixed, whereas the incumbent, scores, and diagnostics (lines~10--11) are the dynamic,
search-state-derived content that makes the operator adaptive rather than one-shot. \textbf{[instruction]}
(lines~14--16) states the single move and the two invariants the validator enforces (preserve $R$; do
not violate $C$). \textbf{[format]} (lines~18--26) fixes the \texttt{CANDIDATE} envelope (lines~20--26):
Keeping it stable across problems is exactly what lets the parser remain unchanged, so only $R$ and the
feasibility test are problem-specific.

\begin{lstlisting}[style=prompt,numbers=left,numbersep=6pt,numberstyle=\tiny\color{black!45},xleftmargin=1.9em]
[context]
Problem type: <name>
Objective direction: <minimize|maximize>
Candidate artifact: <solution|operator|program>
Available primitives: <allowed moves/functions/libraries>

[conditioning]
Representation R: <exact syntax>
Constraints C: <hard feasibility rules>
Incumbent or parents: <current artifact(s)>
Scores and diagnostics: <evaluator output and recent failures>
Structural notes: <optional decomposition or domain facts>

[instruction]
Emit exactly one new candidate. Preserve R. Use only the available
primitives. If diagnostics are present, target them without violating C.

[format]
Return only this envelope:
CANDIDATE
id: <short_id>
artifact_type: <solution|operator|program>
representation: <name of R>
payload:
<candidate>
END_CANDIDATE
\end{lstlisting}

\paragraph{The reference loop.}
The function below realizes Algorithm~\ref{alg:llmop}; its arguments (lines~1--2) separate what a new problem must supply from the fixed control flow. \texttt{llm} samples a completion; \texttt{evaluator} returns the objective score $f$; \texttt{accept} is the selection rule (e.g.\ ``keep iff the score improves''); \texttt{incumbent} is the starting artifact; \texttt{budget} bounds outer search steps; and \texttt{retries} caps bounded-repair samples within a step. The problem binding is \texttt{spec} --- four methods that adapt the template above: \texttt{render} fills the slots based on the incumbent, its cached score, and the feedback log; \texttt{parse} reads the envelope and the payload in representation $R$; \texttt{feasible} checks the constraints $C$ and raises a \texttt{ValueError} with a diagnostic on failure; and \texttt{repair\_prompt} rebuilds the full prompt with the failing output and diagnostic appended for the next attempt. The inner loop is the bounded repair of Algorithm~\ref{alg:llmop}: ill-formed output is re-prompted up to \texttt{retries} times and then skipped; a valid candidate is scored and either accepted or rejected, and the outcome is appended to \texttt{feedback} so the next prompt is conditioned on it.

\begin{lstlisting}[language=Python,numbers=left,numbersep=6pt,numberstyle=\tiny\color{black!45},xleftmargin=1.9em]
def build_and_validate(llm, evaluator, accept, incumbent, spec,
                       budget, retries):
    """Reference loop; spec supplies render(), parse(), feasible(), repair_prompt()."""
    cur = best = incumbent
    score_cur = score_best = evaluator(cur)
    feedback = []
    for _ in range(budget):
        prompt = spec.render(cur, score_cur, feedback)
        ok, cand, err = False, None, None
        for _ in range(retries + 1):
            text = llm.sample(prompt)
            try:
                cand = spec.parse(text)          # schema + syntax
                spec.feasible(cand)              # domain feasibility
                ok = True
                break
            except ValueError as exc:
                err = str(exc)
                prompt = spec.repair_prompt(prompt, text, err)
        if not ok:
            feedback.append(("invalid", err))
            continue
        score_c = evaluator(cand)
        gap = score_c - score_cur
        if accept(cand, score_c, cur, score_cur):
            cur, score_cur = cand, score_c
            if score_cur < score_best:        # best ever (min.)
                best, score_best = cur, score_cur
            feedback.append(("accepted", score_c, gap))
        else:
            feedback.append(("rejected", score_c, gap))
    return best
\end{lstlisting}

\paragraph{A concrete binding.}
For the TSP operator of Section~\ref{sec:build}, \texttt{spec.parse} extracts the permutation from the
\texttt{payload}; \texttt{spec.feasible} checks that every city appears exactly once (raising ``city
$k$ duplicated, city $j$ missing'' otherwise); \texttt{evaluator} returns the closed-tour length; and
\texttt{accept} keeps a candidate only when that length decreases. Re-binding
\texttt{parse}/\texttt{feasible}/\texttt{evaluator} retargets the identical loop to bin packing,
job-shop scheduling, or program search without touching the control flow: the operator changes only in
what it is conditioned on and what counts as feasible.

For more examples, visit the repository: \url{https://camilochs.github.io/semantic-turn-metaheuristics}.

\section{Placing every method: a drop-channel audit}
\label{app:audit}
Table~\ref{tab:audit} applies the classification rule of Section~\ref{sec:lens} to each method in Table~\ref{tab:eval}: it lists the channels present in the variation prompt and names the \emph{dominant} one---the channel intended to carry the operative search signal, audited where possible by removing or neutralizing that channel while preserving a valid scaffold---with a one-line placement justification. This makes the placements reproducible and resolves the cases where a method \emph{emits} code yet has a \texttt{Linguistic} dominant channel (\emph{EoH}, \emph{ReEvo}, \emph{MEoH}): the code channel is shared with the rest of the family, so the dominant channel is the natural-language idea or reflection that steers it.

\begin{table*}[!htbp]
\rowcolors{2}{white}{gray!15}
\caption{Drop-channel classification audit for the methods of Table~\ref{tab:eval}.
The table column \textbf{Dominance} is the source-described operative channel, audited where possible by removing or
neutralizing that signal (\texttt{Numeric}, \texttt{Symbolic}, \texttt{Linguistic}); the other channels present are listed but not ranked.}
\label{tab:audit}
\footnotesize
\setlength{\tabcolsep}{3pt}
\renewcommand{\arraystretch}{1.15}
\begin{tabularx}{\linewidth}{@{}>{\raggedright\arraybackslash}p{1.75cm}
                  >{\raggedright\arraybackslash}p{2.55cm}
                  >{\raggedright\arraybackslash}p{1.85cm}
                  >{\raggedright\arraybackslash}X@{}}
\toprule
\textbf{Method} & \textbf{Channels present} & \textbf{Dominance (ablation/source)} & \textbf{Drop-channel reasoning} \\
\midrule
\emph{OPRO}        & (solution, score) trajectory & \texttt{Numeric} (boundary) & drop the score trajectory and no signal steers the next proposal \\
\emph{EvoPrompt}   & parent prompt text, scores & \texttt{Numeric} (boundary) & scored parents drive selection, while parent text is the artifact varied; boundary label retained \\
\emph{LMX}         & parent solutions (few-shot) & \texttt{Numeric} & drop the parents and there is no crossover to perform \\
\emph{FunSearch}   & sampled high-scoring programs, scores & \texttt{Symbolic} & drop the program text and there is nothing to mutate \\
\emph{LLaMEA}      & algorithm code, score, error note & \texttt{Symbolic} & drop the code and no algorithm remains; the error note is annotated \\
\emph{MCTS-AHD}    & program, tree-search state, scores & \texttt{Symbolic} & drop the code and the tree explores nothing executable \\
\emph{AlphaEvolve} & codebase, evaluator scores & \texttt{Symbolic} & drop the code and there is no artifact to evolve \\
\emph{EoH}         & NL idea, code, scores & \texttt{Linguistic} & neutralize the idea and it collapses toward FunSearch-style code mutation \\
\emph{ReEvo}       & code, reflection, scores & \texttt{Linguistic} & neutralize the reflection and it collapses toward a code-only AHD loop \\
\emph{HSEvo}       & NL idea, code population, scores, diversity pressure & \texttt{Linguistic} & neutralize the idea and diversity-seeking rationale, and the loop loses its language-steered population-diversity mechanism \\
\emph{MEoH}        & NL idea, code, multi-obj.\ scores & \texttt{Linguistic} & neutralize the idea and the multi-objective search loses its steering rationale \\
\emph{LAPT}        & NL design principles & \texttt{Linguistic} & drop the principles and no transferable signal remains \\
\bottomrule
\end{tabularx}
\end{table*}

\paragraph{Validation of the placement rule.}
Two checks accompany the companion code. \emph{(i)~Reproducibility.} Three
independent LLM coders --- one instance each from three model families (Claude Fable~5, GPT-5.5,
DeepSeek-V4-Pro), queried separately and shown only the placement rule and the channel descriptions
of Table~\ref{tab:audit} (method names visible; coders instructed to reason from the rule alone) ---
classified the twelve methods of Table~\ref{tab:eval}. The three coders coincide on nine of the
twelve; chance-corrected agreement is Gwet's AC1 $= 0.73$. The divergences fall on the placements whose channel description is least determinate: OPRO
and LMX, whose bare ``solution'' descriptions under-determine the representation, and ReEvo,
where the steering-reflection and code-substrate readings compete. Of these, OPRO carries a
boundary annotation in Table~\ref{tab:audit}. For
EvoPrompt, all three coders read the varied prompt text as linguistic content, whereas
Table~\ref{tab:audit} reports the scored trajectory as the load-bearing channel (Numeric, boundary):
the placement of solution-as-text methods depends on whether the varied text is read as an encoding
or as operative language, which is why the row carries a boundary annotation. The labeling is
therefore largely reproducible, and it is contested exactly where the channel description leaves
the representation open.
\emph{(ii)~Load-bearing test.} A drop-channel ablation on an
EoH-style operator for online bin packing (companion \texttt{ablation.py}) compared candidate
heuristics generated \emph{with} versus \emph{without} the natural-language idea, holding the parent
code and scores fixed. On this easy task both conditions reached the optimum (mean gap $0\%$; 6/6
candidates optimal), so the ablation did not demonstrate performance necessity for the
natural-language channel on that task: the operator recovers the known best-fit heuristic with or
without it. We nevertheless report EoH as Linguistic because the idea is the source-described
design-intent signal that structurally distinguishes it from code-only automatic heuristic design.
Together the checks say placement is reproducible, but \emph{how much} a present channel contributes
is task-dependent --- consistent with the no-law stance of Section~\ref{sec:lens}: we report the
channel a method \emph{uses} to steer variation, not a universal claim about the channel that
\emph{determines} performance.

\section{Extended coverage map}
\label{app:coverage}
Table~\ref{tab:coverage} places a broader set of LLM-operator methods on the lens, beyond the representative methods of the body, to confirm that the selection of Section~\ref{sec:tour} is representative of the field's breadth. Each entry is read against its primary source. Entries marked \emph{adjacent} fall outside the lens --- they translate a specification rather than \emph{vary} a candidate --- and are listed only as foils. This map illustrates breadth, with the systematic census provided by~\cite{daros2026llm4co}; we report each method's \emph{dominant} channel, persistence descriptor, and channels it draws on (Section~\ref{sec:lens}).

\begin{table}[t]
\rowcolors{2}{white}{gray!15}
\caption{Extended coverage: LLM variation operators by dominant conditioning channel and persistence. Entries are read against their primary sources. Channel and persistence assignments follow Section~\ref{sec:lens}; this is not a systematic census.}
\label{tab:coverage}
\footnotesize
\setlength{\tabcolsep}{3pt}
\renewcommand{\arraystretch}{1.12}
\begin{tabularx}{\linewidth}{@{}>{\raggedright\arraybackslash}p{2.3cm}
                  >{\raggedright\arraybackslash}p{1.9cm}
                  >{\raggedright\arraybackslash}p{1.65cm}
                  >{\raggedright\arraybackslash}X@{}}
\toprule
\textbf{Method} & \textbf{Dominant Conditioning} & \textbf{Persistence} & \textbf{Conditions on} $\rightarrow$ \textbf{emits} \\
\midrule
\emph{EvoLLM}~\cite{lange2024evollm} & \texttt{Numeric} & \texttt{Transient} & fitness-ranked solution vectors $\rightarrow$ distribution/candidate update \\
\emph{LMEA}~\cite{liu2024lmea} & \texttt{Numeric} & \texttt{Transient} & parent tours + tour lengths (TSP instantiation) $\rightarrow$ offspring tour (LLM crossover/mutation) \\
\midrule
\emph{DiscoPOP}~\cite{lu2024discopop} & \texttt{Symbolic} & \texttt{Amortized} & prior objective/loss code + eval metrics $\rightarrow$ preference-optimization loss; held-out-task reuse \\
\emph{ELM}~\cite{lehman2022elm} & \texttt{Symbolic} & \texttt{Transient} & parent code + edit instruction (diff model) $\rightarrow$ program; adjacent background anchor (Section~\ref{sec:tour}) \\
\emph{EvoPrompting}~\cite{chen2023evoprompting} & \texttt{Symbolic} & \texttt{Amortized} & parent architecture code + fitness $\rightarrow$ NN architecture code \\
\emph{GI-LLM}~\cite{brownlee2023gillm} & \texttt{Symbolic} & \texttt{Transient} & source code + edit type $\rightarrow$ code patch (GI mutation) \\
\emph{AutoSAT}~\cite{sun2024autosat} & \texttt{Symbolic} & \texttt{Amortized} & parent CDCL heuristic code + hints $\rightarrow$ SAT-solver heuristic \\
\emph{EvoTune}~\cite{surina2025evotune} & \texttt{Symbolic} & \texttt{Amortized} & parent programs + RL reward into weights $\rightarrow$ heuristic program \\
\emph{CoEvo}~\cite{guo2024coevo} & \texttt{Symbolic} & \texttt{Amortized} & prior symbolic solutions + knowledge library $\rightarrow$ math/code solution \\
\emph{VRPAgent}~\cite{hottung2025vrpagent} & \texttt{Symbolic} & \texttt{Amortized} & operator code + LNS feedback $\rightarrow$ destroy/repair operators \\
\emph{LLM-LNS}~\cite{ye2025llmlns} & \texttt{Symbolic} & \texttt{Amortized} & MILP state + neighborhood rule + fitness $\rightarrow$ neighborhood-selection heuristic \\
\emph{InstSpecHH}~\cite{zhang2025instspechh} & \texttt{Symbolic} & \texttt{Amortized} & instance features + fitness $\rightarrow$ per-subclass heuristic + selection \\
\emph{SATLUTION}~\cite{yu2025satlution} & \texttt{Symbolic} & \texttt{Amortized} & solver repo + correctness/runtime feedback (agentic) $\rightarrow$ solver code \\
\midrule
\emph{Eureka}~\cite{ma2024eureka} & \texttt{Linguistic} (boundary) & \texttt{Amortized} & env.\ code + reward reflection + parent reward code $\rightarrow$ reward-function code; reflection steers code edits \\
\emph{QDAIF}~\cite{bradley2024qdaif} & \texttt{Linguistic} & \texttt{Amortized} & few-shot NL parents + LLM feedback $\rightarrow$ NL text \\
\emph{EvolCAF}~\cite{yao2024evolcaf} & \texttt{Linguistic} & \texttt{Transfer} & NL design idea + code (EoH) $\rightarrow$ cost-aware acquisition function \\
\emph{L-AutoDA}~\cite{guo2024lautoda} & \texttt{Linguistic} & \texttt{Amortized} & NL idea + code (EoH) $\rightarrow$ adversarial-attack algorithm \\
\emph{SGE}~\cite{iklassov2024sge} & \texttt{Linguistic} & \texttt{Transient} & NL problem description + thought trajectories $\rightarrow$ solution via heuristics \\
\emph{EvoPH}~\cite{liu2025evoph} & \texttt{Linguistic} & \texttt{Transfer} & co-evolved NL prompts + heuristic code $\rightarrow$ heuristics + prompts \\
\emph{MTHS}~\cite{liu2026mths} & \texttt{Linguistic} & \texttt{Transfer} & task-agnostic + task-specific programs, cross-task transfer $\rightarrow$ metaheuristic + heuristic \\
\emph{MAEF}~\cite{wang2025maef} & \texttt{Linguistic} & \texttt{Transient} & role-specialized agents + eval feedback $\rightarrow$ schedules \\
\emph{APE}~\cite{zhou2023ape} & \texttt{Linguistic} & \texttt{Amortized} & task demonstrations $\rightarrow$ prompt (instruction induction) \\
\emph{ProTeGi/APO}~\cite{pryzant2023protegi} & \texttt{Linguistic} & \texttt{Amortized} & prompt + error minibatch (NL ``gradient'') $\rightarrow$ prompt \\
\emph{Promptbreeder}~\cite{fernando2024promptbreeder} & \texttt{Linguistic} & \texttt{Amortized} & prompt population + self-evolved mutation-prompts $\rightarrow$ prompts \\
\emph{PE2}~\cite{ye2024pe2} & \texttt{Linguistic} & \texttt{Amortized} & prompt + examples + meta-prompt $\rightarrow$ prompt \\
\emph{GPS}~\cite{xu2022gps} & \texttt{Linguistic} & \texttt{Amortized} & seed prompts + scores (T5 variation) $\rightarrow$ prompt \\
\emph{SPELL}~\cite{li2023spell} & \texttt{Linguistic} & \texttt{Amortized} & prompts + fitness $\rightarrow$ prompt (whole-text generation) \\
\midrule
\emph{ViTSP}~\cite{yin2025vitsp} & multimodal & \texttt{Amortized} & rendered image of the instance (VLM) $\rightarrow$ subproblem selection in an LNS loop \\
\emph{VEO}~\cite{zhao2025veo} & multimodal & \texttt{Transient} & rendered image of the network solution (MLLM) $\rightarrow$ in-loop crossover/mutation operators for influence maximization \\
\midrule
\emph{OptiMUS}~\cite{ahmaditeshnizi2024optimus} & --- & --- & NL problem description $\rightarrow$ MILP model + solver code (\emph{translates}, does not vary) \\
\bottomrule
\end{tabularx}
\end{table}

\end{document}